\documentclass{article} % For LaTeX2e
\usepackage{iclr2025_conference,times}

\usepackage{amsmath,amsfonts,bm}

\def\eqref#1{equation~\ref{#1}}
\def\1{\bm{1}}

\def\mQ{{\bm{Q}}}

\def\mV{{\bm{V}}}

\DeclareMathAlphabet{\mathsfit}{\encodingdefault}{\sfdefault}{m}{sl}
\SetMathAlphabet{\mathsfit}{bold}{\encodingdefault}{\sfdefault}{bx}{n}

\usepackage{hyperref}
\usepackage{url}

\usepackage{macros}

\title{Learning Efficient Positional Encodings with Graph Neural Networks}

\author{%
  Charilaos I.~Kanatsoulis\\
  Dept. of Computer Science\\
  Stanford University\\
  \texttt{charilaos@cs.stanford.edu} \\
  \And
    Evelyn~Choi\\
  Dept. of Computer Science\\
  Stanford University\\
  \texttt{echoi1@stanford.edu} \\
  \And
Stephanie Jegelka\\
School of Computation, Information and Technology\\
Technical University of Munich\\
Computer Science \& Artificial Intelligence Laboratory\\
Massachusetts Institute of Technology\\
  \texttt{stefje@mit.edu} \\
  \And
Jure Leskovec$^*$\\
  Dept. of Computer Science\\
  Stanford University\\
  \texttt{jure@cs.stanford.edu} \\
   \And
    Alejandro Ribeiro\thanks{Equal senior authorship.}\\
  Dept. of Electrical and Systems Engineering\\
  University of Pennsylvania\\
  \texttt{aribeiro@seas.upenn.edu} \\
}

\iclrfinalcopy % Uncomment for camera-ready version, but NOT for submission.
\begin{document}

\maketitle
\begin{abstract}
Positional encodings (PEs) are essential for effective graph representation learning because they provide position awareness in inherently position-agnostic transformer architectures and increase the expressive capacity of Graph Neural Networks (GNNs). However, designing powerful and efficient PEs for graphs poses significant challenges due to the absence of canonical node ordering and the scale of the graph. {In this work, we identify four key properties that graph PEs should satisfy}: stability, expressive power, scalability, and genericness. We find that existing eigenvector-based PE methods often fall short of jointly satisfying these criteria. To address this gap, we introduce \method, a novel framework of learnable PEs for graphs. Our primary insight is that message-passing GNNs function as nonlinear mappings of eigenvectors, enabling the design of GNN architectures for generating powerful and efficient PEs. A crucial challenge lies in initializing node attributes in a manner that is both expressive and permutation equivariant. We tackle this by initializing GNNs with random node inputs or standard basis vectors, thereby unlocking the expressive power of message-passing operations, while employing statistical pooling functions to maintain permutation equivariance. Our analysis demonstrates that \method~approximates equivariant functions of eigenvectors with linear complexity, while rigorously establishing its stability and high expressive power. Experimental evaluations show that \method~outperforms lightweight versions of eigenvector-based PEs and achieves comparable performance to full eigenvector-based PEs, but with one or two orders of magnitude lower complexity. Our
code is available at~\url{https://github.com/ehejin/Pearl-PE}.
\end{abstract}

\section{Introduction}
Positional encodings (PEs) are a fundamental component of graph representation learning and play a key role in the design of effective Graph Transformers \citep{dwivedi2020generalization,rampavsek2022recipe} and Graph Neural Networks (GNNs) \citep{kipf2016semi,hu2020strategies}. Transformer architectures \citep{vaswani2017attention} are inherently agnostic to structure and node identities, and PEs provide a powerful mechanism to incorporate positional and structural information. On the other hand, message-passing GNNs often struggle with low expressiveness, especially when node attributes exhibit the same symmetries as the graph structure \citep{xu2018powerful,morris2019weisfeiler,kanatsoulis2024graph}. By integrating structural and positional information, PEs enhance GNNs' capacity to capture patterns that would otherwise be difficult to learn and generalize.

Several graph PE methods have been proposed in the literature, which can broadly be categorized into two main types: absolute PEs and relative PEs. Absolute PEs assign an embedding to each node in the graph, reflecting the node's role within the graph structure. Common approaches include Laplacian eigenvectors \citep{dwivedi2020generalization}, substructure encodings \citep{tahmasebi2020counting, you2021identity, bouritsas2022improving}, random walk (RW) encodings \citep{rampavsek2022recipe}, and eigenvector-based methods \citep{kreuzer2021rethinking, limsign, huangstability}. Relative PEs, on the other hand, assign representations to pairs of nodes and typically utilize measures such as shortest-path and resistance distances \citep{ying2021transformers, zhang2023rethinking}, as well as RW matrices \citep{ma2023graph,geisler2023transformers}. A thorough comparison between absolute and relative PEs can be found in \citep{black2024comparing}.

In this paper, we study absolute PEs for graphs based on four key criteria: expressive power, scalability, stability under perturbations, and generality. We find that PEs based on eigenvectors of graph Laplacian or other graph operators often struggle to satisfy all these criteria simultaneously. To better understand this, we divide eigenvector-based approaches into two categories: those that compute the full set of eigenvectors and those that only consider the \(K\) largest. Full eigenvector approaches offer high expressive power but come with a computational complexity of \(\mathcal{O}(N^3)\) and memory complexity of \(\mathcal{O}(N^2)\), which is prohibitive for even medium-sized graphs. The full set of eigenvectors can also be used to learn spectral graph filters \citep{huangstability}, which result in stable PEs. Note that stability is particularly crucial for out-of-distribution generalization.

However, when only a subset of eigenvectors is computed, several limitations arise. First, this introduces an inductive bias, as different graphs encode different information across eigenvalues, especially when they differ in size. Second, the expressive power and stability are reduced, becoming dependent on the eigengap between the selected eigenvalues. These methods also face challenges in terms of stability and generalization when applied to different or unseen graph structures. Consequently, this approach often leads to significantly poorer performance. Substructure-based encodings face similar challenges: while generally stable, they also introduce inductive bias and highly expressive versions require combinatorial complexity. The aforementioned challenges raise a critical research question:

\textbf{Question:} Can we learn generic PEs that are simultaneously expressive, stable, and scalable?

In this work, we provide an affirmative answer by proposing \method, a powerful and efficient framework for learnable PEs, entirely generated via message-passing GNNs. We begin by showing that message-passing GNNs can be understood as nonlinear mappings of eigenvectors of the graph Laplacian or other graph shift operators. This insight enables the computation of eigenvector-based PEs efficiently with linear or quadratic complexity, leveraging message-passing operations. A central challenge in developing effective PEs with GNNs lies in initializing node attributes to ensure both expressiveness and permutation equivariance. We address this by initializing each node with a set of \(M\) random samples, effectively breaking the symmetries between the graph structure and node attributes. Each sample is processed independently by a GNN, and to guarantee permutation equivariance, we design pooling functions based on statistics. Our analysis demonstrates that \method~surpasses the expressiveness of the Weisfeiler-Leman (WL) test \citep{weisfeiler1968reduction}, and is capable of counting key substructures at the node level.

\method~is provably stable, inheriting the stability guarantees of GNNs \citep{gama2020stability}, which are independent of the eigenvalue gap. Moreover, we analyze the sample complexity of \method~and show that the number of samples required for effective encoding is independent of graph size. This enables the generation of powerful eigenvector-based PEs for large graphs with linear complexity. For smaller graphs, where the number of samples is comparable to the graph size, we propose an alternative model that initializes node attributes with basis vectors. This approach approximates the PEs in \citep{huangstability} with significantly lower computational and memory complexity. We evaluate the proposed \method~on graph classification and regression tasks on molecular graphs and social network datasets, and compare it against eigen-based and structure-based absolute PEs. The results demonstrate that \method~consistently outperforms structure-based PEs and lightweight variants of eigenvector-based PEs, achieving up to a $6\%$ improvement on graph classification tasks. In comparison to full eigenvector-based PEs, which have a computational complexity of $\mathcal{O}(N^3)$, \method~delivers comparable performance with significantly reduced complexity, scaling at $\mathcal{O}(N)$ or $\mathcal{O}(N^2)$.

\section{Preliminaries}
A graph $\mathcal{G}:=(\mathcal{V},\mathcal{E})$, is represented by a set of vertices $\mathcal{V}=\{1,\dots,N\}$, a set of edges $\mathcal{E}=\left\{\left(v,u\right)\right\}$, and a graph shift operator (GSO) $\bm S\in\mathbb{R}^{N\times N}$. The GSO is typically sparse, with common choices including the adjacency matrix, the Laplacian matrix, their normalized variants, or the RW transition matrix. The nodes (vertices) in the graph are often associated with node signals \(\bm{x}_v \in \mathbb{R}^{d}\), each with \(d\) features, while edges can carry edge attributes \(\bm{x}_{(u,v)} \in \mathbb{R}^{d_e}\) with \(d_e\) features.

An important operation in graph theory and network science is the spectral decomposition of the graph and refers to the eigenvalue decomposition to the GSO, $\bm S = \bm V\bm\Lambda\bm V^T$. Matrix $\bm V = \left[\bm v_1,\dots,\bm v_n\right]$ is the orthonormal matrix of eigenvectors, and $\bm \Lambda$ is the diagonal matrix of eigenvalues $\left\{\lambda_n\right\}_{n=1}^N$. When $\bm S$ represents the Laplacian matrix, $\bm V$ are the Laplacian eigenvectors that are commonly used as node features or positional encodings for GNN architectures.

In this paper, we study standard message-passing GNNs, defined by the following recursive formula:
\begin{align}\label{eq:GNNrec00}
   \bm x_v^{(l)} = g^{(l-1)}\left(\bm x_v^{(l-1)},f^{(l-1)}\left(\left\{\bm x_u^{(l-1)}:u\in\mathcal{N}\left(v\right)\right\}\right)\right).
\end{align}
Here, $\mathcal{N}(v)$ represents the neighborhood of vertex $v$, meaning that $u \in \mathcal{N}(v)$ if and only if $(u, v) \in \mathcal{E}$. The function $f^{(l)}$ aggregates information from the multiset of signals coming from neighboring vertices, while $g^{(l)}$ combines the signal of each vertex with the aggregated information from its neighbors. Common choices for $f^{(l)}$ and $g^{(l)}$ include the single- and multi-layer perceptron (MLP), the linear function, and the summation function.
\section{Our Work: Learnable, Efficient, and Powerful PEs with GNNs}
\subsection{GNNs are nonlinear functions of GSO eigenvectors}
Our first observation is that message-passing GNNs are nonlinear functions of eigenvectors. To see this, let $f^{(l)}$ be one of the following aggregation functions:
\begin{align}\label{eq:agg_functions}
\sum_{u\in\mathcal{N}\left(v\right)}\bm x_u,~d_v\cdot\bm x_v-\sum_{u\in\mathcal{N}\left(v\right)}\bm x_u,~\sum_{u\in\mathcal{N}\left(v\right)}\frac{\bm x_u}{\sqrt{d_v d_u}},~\bm x_u-\sum_{u\in\mathcal{N}\left(v\right)}\frac{\bm x_u}{\sqrt{d_v d_u}},~\sum_{u\in\mathcal{N}\left(v\right)}\frac{\bm x_u}{{d_u}}
\end{align}
where $d_v$ is the degree of node $v$. Then Eq. (\ref{eq:GNNrec00}) can be written as $\bm X^{(l)} = g^{(l-1)}\left(\bm X^{(l-1)},\bm S\bm X^{(l-1)}\bm\right)$, where $\bm S$ represents the adjacency matrix, the Laplacian matrix, the normalized adjacency, the normalized Laplacian, and the RW matrix, for the five choices of $f^{(l)}$ in \ref{eq:agg_functions} respectively, and $\bm X^{(l)}\in\mathbb{R}^{N\times F_{l}}$ represents the signals of all vertices at layer $l$. Now let $g^{(l)}$ be an equivariant MLP operating on each node independently. Note that the MLP is a common choice for function $g^{(l)}$ for the majority of effective GNN architectures due to its expressiveness properties. Then Eq. (\ref{eq:GNNrec00}) can be cast as:
\begin{equation}\label{eq:GNNrec02}
    \bm X^{(l)} =  \sigma\left(\sum_{k=0}^{K-1}\bm S^k \bm X^{(l-1)}\bm H_k^{(l)}\right),
\end{equation}
where $K=2$, $\bm H_k\in\mathbb{R}^{F_{l-1}\times F_{l}}$ are the trainable parameters, and $\sigma$ is a point-wise nonlinear activation function. Note that Eq. (\ref{eq:GNNrec02}) defines a single-layer graph perceptron, but it can be easily generalized to a multi-layer graph perceptron by letting $\sigma$ represent an equivariant MLP acting on the node signals. Additionally, while we set $K=2$ here, higher values of $K$ can be considered for more generalized GNN layers. It is worth emphasizing that $\bm S^k$ is never explicitly instantiated; instead, $\bm S^k \bm X^{(l-1)}$ is computed using recursive message-passing operations, as outlined in Eq. (\ref{eq:agg_functions}).

\begin{proposition}[GNNs are nonlinear functions of eigenvectors]\label{prop:eig}
A GNN defined in Eq. (\ref{eq:GNNrec00}) with $f^{(l)}$ being one of the functions in Eq. (\ref{eq:agg_functions}) and $g^{(l)}$ being a multi-layer perceptron, {operates as} a nonlinear function of the GSO eigenvectors i.e., {$\bm x_v^{(l)} =\texttt{MLP}\left(\bm v^{(v)}\right),~\bm v^{(v)}=\bm V[v,:]^T$}. The trainable parameters of the first \texttt{MLP} layer are not independent but depend on the eigenvalues $\left\{\lambda_n\right\}_{n=1}^N$ and eigenvectors $\left\{\bm v_n\right\}_{n=1}^N$ of the GSO, as well as the node features $\bm X$ of the graph:
{\begin{equation}
\bm x_v^{(l)} =\texttt{MLP}\left(\bm v^{(v)}\right)=\texttt{MLP}^{(-1)}\left(\sigma\left(\bm W^T\bm v^{(v)}\right)\right)
\end{equation}}
\begin{equation}\label{eq:dof}
\bm W\left[n, f\right] = \sum_{i=1}^{F_{l-1}}\sum_{k=0}^{K-1}\lambda_n^k\bm H_k^{(l)}[i,f]\langle {\bm \alpha_n}, \bm X^{(l-1)}\left[:,i\right]\rangle,
\end{equation}
{where $\bm\alpha_n = \bm v_n$ when the GSO is symmetric and $\bm\alpha_n = \bm V^{-1}[n,:]$ when it is not.} $\texttt{MLP}^{(-1)}$ denotes all the layers of the $\texttt{MLP}$ except the first layer. 
\end{proposition}
The proof is provided in Appendix \ref{app:MLP}. {According to Proposition \ref{prop:eig}, the update for node \( v \), defined by the function \( g \circ f : \left(\mathcal{G},\mathbb{R}^{F_{l-1}}\right) \rightarrow \mathbb{R}^{F_l} \), can be interpreted as a nonlinear mapping (\texttt{MLP}) applied to \( \bm V[v, :] \), but the weights of the first layer of this mapping are also mappings, i.e., $\mathbf{W}[n,f] : \left(\mathcal{G},\mathbb{R}^{F_{l-1}}\right)\rightarrow \mathbb{R}^{1}$. The degrees of freedom in the first layer of \( \texttt{MLP} \) are \( K F_l F_{l-1} \) (as described in Eq. (\ref{eq:dof})), rather than \( F_l N \), which would be the case for independent weights \( \bm W \). Furthermore, the dot product \( \langle \bm \alpha_n, \bm X^{(l-1)}[:, i] \rangle \) depends on the eigenvectors and, for the update of node \( v \), it only involves the components \( \bm X^{(l-1)}[u, i],~u\in\mathcal{N}_v \).} Proposition \ref{prop:eig} is applicable to most message-passing GNN models, including, but not limited to, Graph Convolutional Networks (GCNs) \citep{kipf2016semi}, Graph Isomorphism Networks (GINs) \citep{xu2018powerful}, and GraphSAGE \citep{hamilton2017inductive}.

\subsection{\method: Expressive and Equivariant Positional Encoding Networks}
\begin{figure}[t]
\begin{center}
%\framebox[4.0in]{$\;$}
% \fbox{\rule[-.5cm]{0cm}{4cm} \rule[-.5cm]{4cm}{0cm}}
\includegraphics[width=1\linewidth]{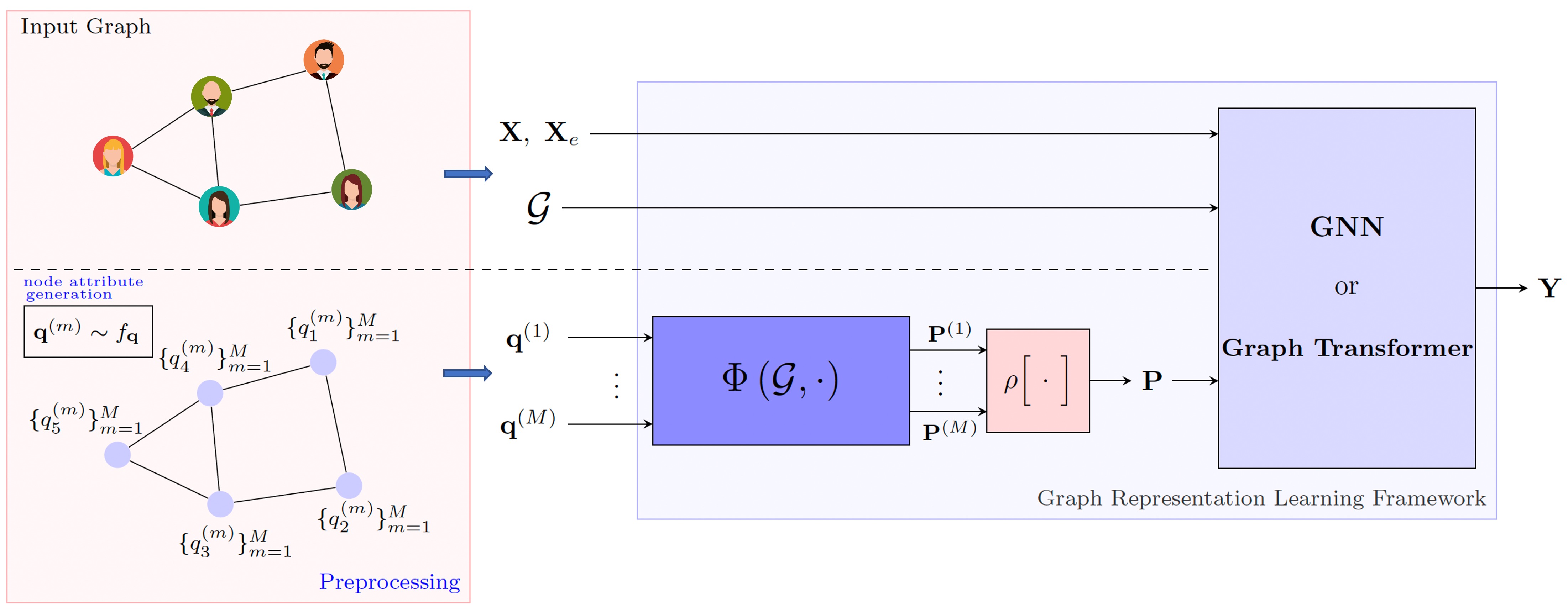}
\end{center}
\caption{\method~framework: The input graph undergoes anonymization by removing its node and edge attributes. For each node, a set of $M$ random or basis attributes is generated. Each sample is then independently processed by a message-passing GNN, and a pooling function $\rho$ is applied to produce equivariant PEs. The graph structure, together with the generated PEs and any node or graph attributes, is subsequently processed using either a GNN or a Graph Transformer.
}
\label{fig:PEframework}
\end{figure}
Following the derivation of Proposition \ref{prop:eig}, a critical question arises: what is the optimal choice of node attributes that allow a GNN to compute expressive and equivariant functions of the eigenvectors? Equivariant structural features augment GNNs with valuable information, but they come at the cost of increased computational complexity and inductive bias. Moreover, these features share the same symmetries as the graph structure, limiting the expressiveness of message-passing GNNs. Alternatively, unique identifiers, such as random node features, can break the structural symmetries and improve expressiveness but at the expense of permutation equivariance, which limits the model’s generalization capability. To address this trade-off between equivariance and expressiveness, we propose to momentarily break the structural symmetries by initializing each node with a set of $M$ unique identifiers, while maintaining permutation equivariance in the model output via the use of statistical pooling functions. The proposed PE framework (\method) is illustrated in Fig. \ref{fig:PEframework} and, as we see next, ensures both high expressiveness and strong generalization.

Consider a graph $\mathcal{G}=(\mathcal{V},\mathcal{E})$ with $N$ nodes. For each node $v\in\mathcal{V}$ in $\mathcal{G}$, we design a set of $M$ $1-$dimensional node signals $\left\{  q_v^{(1)}, q_v^{(2)},\dots,  q_v^{(M)} \right\}$, where each $q_v^{(m)}$ operates as a unique identifier. Graph $\mathcal{G}$ is now associated with a set of $M$ independent initial node attributes represented as \(\left\{  \bm q^{(m)}\right\}_{m=1}^M\), $\bm q^{(m)}\in\mathbb{R}^N$. Each pair of $\left\{\mathcal{G},\bm q^{(m)}\right\}$ is independently processed via a GNN \(\Phi(\cdot)\), which is described by Eq. (\ref{eq:GNNrec00}) or (\ref{eq:GNNrec02}), to produce a set of $M$ independent outputs:
\begin{equation}
     {\bm P}^{(m)} = \Phi\left( \mathcal{G},\bm q^{(m)} \right) \in \mathbb{R}^{N \times d_p },\quad m= 1,\dots, M
\end{equation}
Since \(\left\{ \bm{q}^{(m)} \right\}_{m=1}^M\) operate as unique identifiers, they break the structural symmetries and unlock the expressive power of message-passing operations. However, each \({\bm P}^{(m)}\) is not permutation equivariant, thus not generalizable. {To address this, we design an equivariant function \(\rho\), that involves a pooling operation over the independent outputs \(\{\bm{P}^{(m)}\}_{m=1}^M\), to generate the final PE for each node:}
\begin{equation}
        \bm P = \rho\left[ \Phi\left( \mathcal{G},\bm q^{(1)} \right),\dots,\Phi\left( \mathcal{G},\bm q^{(M)} \right) \right]\in \mathbb{R}^{N \times d_p}
\end{equation}
{The \method~framework can universally approximate any continuous basis invariant function.}
{\begin{theorem}[Basis Universality]\label{theorem: SPE express BasisNet}
    Let $\mathcal{G}$ be a graph with GSO $\bm S = \bm V\bm\Lambda\bm V^T$, and $f$ be a continuous function such that $f(\mV) = f(\mV \mQ)$, $\mQ \in \mathcal{O}\left(\bm \diag\left({\bm\Lambda}\right)\right)$, for any eigenvalues $\bm {\Lambda}$. Then there exist GNN $\Phi$ and a continuous equivariant function $\rho$, such that $f(\mV) = \rho\left[ \Phi\left( \mathcal{G},\bm q^{(1)} \right),\dots,\Phi\left( \mathcal{G},\bm q^{(M)} \right) \right]$. 
\end{theorem}}
The proof can be found in Appendix \ref{app:universality}. {Theorem \ref{theorem: SPE express BasisNet} can be extended to handle multiple graphs by considering $\mathcal{G}$ as a collection of graphs represented as disconnected components. In this case, the GSO takes a block diagonal form, where each block corresponds to the GSO of an individual graph.} In the following sections, we explore options for the initial node attributes \(\left\{ \bm{q}^{(m)} \right\}_{m=1}^M\) and pooling functions \(\rho\). A key aspect of \method~is designing \(M\) independent initial attributes for each node, which enables permutation equivariance at the model’s output through the use of pooling functions. This stands in contrast to classical methods, which typically assign a single unique identifier per node.

\section{Our work: Random Positional Encoding Network (R-\method)}
Next, we present our Random PE Network (R-\method).
In R-\method~we define node attributes \(\left\{ \bm{q}^{(m)} \right\}_{m=1}^M\) by sampling them randomly from a probability distribution. Specifically, let \(\bm{q} = \left[q_{v_1}, q_{v_2}, \dots, q_{v_N}\right]^T\), where \(v_n \in \mathcal{V}\), be a random vector with joint distribution \(f_{\bm{q}}(t_1, \dots, t_N)\). The set \(\left\{ \bm{q}^{(m)} \right\}_{m=1}^M\) consists of \(M\) independent \(N\)-dimensional realizations of \(\bm{q}\), drawn from \(f_{\bm{q}}\). In our experiments and analysis, \(\bm{q}\) is either a set of independent and identically distributed (i.i.d.) Gaussian random variables or a set of i.i.d. random variables with \(\mathbb{E}\left[q_i\right] = 0\) and \(\mathbb{E}\left[q_i^p\right] = 1\), where \(p \geq 2\).

When these samples are processed by a GNN \(\Phi\), the result is \(M\) \(\left(N \times d_p\right)\)-dimensional samples of the random matrix output \(\Phi(\mathcal{G}, \bm{q})\). To {practically preserve} permutation equivariance, we note that the distribution of \(\Phi(\mathcal{G}, \bm{q})\) is itself permutation equivariant, as are any statistics derived from it. Therefore, the function \(\rho\) can be any empirical statistic computed from the samples \(\left\{\Phi(\mathcal{G}, \bm{q}^{(m)})\right\}_{m=1}^M\), each capturing different characteristics. For instance, \(\rho\) could represent any statistical moment, such as the mean or variance, or other measures as the empirical mode or median. In this paper, we choose \(\rho\) to be the empirical mean due to its favorable convergence and stability properties, as well as its simplicity in implementation and low computational and memory complexity:

\begin{equation}\label{eq:randomPE}
        \bm P = \hat{\mathbb{E}}\left[ \Phi\left( \mathcal{G},\bm q^{(1)} \right),\dots,\Phi\left( \mathcal{G},\bm q^{(M)} \right) \right]=\frac{1}{M}\sum_{m=1}^M\Phi\left( \mathcal{G},\bm q^{(m)} \right)=\frac{1}{M}\sum_{m=1}^M\bm P^{(m)}
\end{equation}
In Appendix \ref{app:taylor}, we explicitly analyze the equivariant functions learned by Eq. (\ref{eq:randomPE}). We derive nonlinear expressions both in the graph domain, using vertex and edge information, and in the frequency domain, using the eigenvectors and eigenvalues of the GSOs. The key to this nonlinear analysis involves studying the pointwise nonlinearities through their Taylor series expansion.
\subsection{Sample Complexity}
In this section, we analyze the number of samples required to such that $\frac{1}{M}\sum_{m=1}^M\bm P^{(m)}$ approximates $\mathbb{E}\left[\Phi\left( \mathcal{G},\bm q \right)\right]$ with negligible error. To that end, we make the following two assumptions.
\vspace{-0.1cm}
\begin{assumption}\label{ass:nonlinearity}
    The pointwise nonlinearity $\sigma$ is Lipschitz continuous with Lipschitz constant \( C_{\sigma} \).
\end{assumption}
\vspace{-0.1cm}
This is a common assumption in deep learning and is satisfied by the widely used nonlinearities. In most cases, such as the Rectified Linear Unit (ReLU), hyperbolic tangent, and sigmoid, it holds that \( C_{\sigma} = 1 \). Before introducing the second assumption, we first need to examine Eq. (\ref{eq:GNNrec02}) more closely. Notice that its linear component involves \( F_{L} \cdot F_{l-1} \) graph filters of the form \( \sum_{k=0}^{K-1} h_k \bm{S}^k \), which is also explicitely shown in Appendix \ref{app:samplecomplexity}, Eq. (\ref{eq:GNNmulti3}).
\vspace{-0.1cm}
\begin{assumption} \label{ass:filter}
    The linear operators $\bm H\left(\bm S\right)=\sum_{k=0}^{K-1}h_k\bm S^k$ involved in the projection of Eq. (\ref{eq:GNNrec02}) are bounded, i.e., $\lVert\bm H\left(\bm S\right)\rVert \leq \beta$.
\end{assumption}
\vspace{-0.1cm}
This is another common assumption in deep learning, where the value of \( \beta \) varies depending on the architecture and task. We can now present Theorem \ref{prop:sample}, which characterizes the number of samples \( M \) needed for our approach to converge to the true \( \mathbb{E}\left[\Phi\left( \mathcal{G},\bm q \right)\right] \).

\begin{theorem}[Sample Complexity]\label{prop:sample}
Let $\bm P$ denote the output of the architecture described in Eq. (\ref{eq:randomPE}), for a graph $\mathcal{G}$ with i.i.d. initial node attributes with unit variance. Also let $\Phi$ be an $L-$layer GNN described by Eq. (\ref{eq:GNNrec02}), with $F$ hidden dimensions at each layer. If $C_{\sigma}=1$ and $\beta=1/F$, the number of samples $M$ required such that:
\begin{equation}
    \left|  \frac{1}{M}\sum_{m=1}^M\Phi\left( \mathcal{G},\bm q^{(m)} \right) - \mathbb{E}\left[\Phi\left( \mathcal{G},\bm q \right)\right] \right| < \epsilon,~\text{with probability at least}~1-\delta,
\end{equation}
satisfies:
\begin{equation}
M \leq \frac{1}{\delta\cdot \epsilon^2}.
\end{equation}
\end{theorem}
It is worth noting that the above bound is independent of the size of the graphs, which suggests that our proposed PE framework is well-suited for large-scale graphs. In practice, we have observed that $10\sim100$ samples are typically sufficient.
\subsection{Expressive Power}
In this section, we establish the expressive power of our proposed R-\method.
\begin{corollary}[Structure Counting]\label{thrm:cycles7}
Let \(\bm{q} = \left[q_1, \dots, q_N\right]\) be a set of \(N\) i.i.d. random variables such that \(\mathbb{E}[q_i] = 0\) and \(\mathbb{E}[q_i^p] = 1\) for \(p \geq 2\). Then, there exists a parametrization \(\Phi\), defined by Eq. (\ref{eq:GNNrec02}), such that \(\mathbb{E}\left[\Phi\left(\cdot, \bm{q}\right)\right]\) counts the number of 3-, 4-, 5-, 6-, and 7-node cycles in which each node participates, for any given graph.
\end{corollary}
Corollary \ref{thrm:cycles7} not only highlights the expressive power of R-\method~framework but also provides valuable insights into its generalization ability. Essentially, R-\method~framework can learn not just the number of cycles each node in a given graph participates in, but also a counting function that generalizes this capability to any node in any graph. The proof can be found in Appendix \ref{app:expr}, and is based on the results in \citep{kanatsouliscounting}. In Corollary \ref{prop:exprpower} we characterize the expressive power of message-passing GNNs with our proposed PEs with respect to the folklore-Weisfeiler-Leman (FWL) test \citep{cai1992optimal,morris2019weisfeiler,huang2021short}.
\begin{corollary}[Expressive Power]\label{prop:exprpower}
A GNN defined in Eq. (\ref{eq:GNNrec00}), with PEs produced by Eq. (\ref{eq:randomPE}) is strictly more powerful than the 1-FWL test, when $f,~g$ are injective functions.
\end{corollary}
The proof of Corollary \ref{prop:exprpower} is a consequence of Corollary \ref{thrm:cycles7} and the analysis in \citep{xu2018powerful}.
Note that the previous results can be improved (e.g., count cycles and cliques of higher order, go beyond 2-FWL test) when the samples \(\left\{ \bm{q}^{(m)} \right\}_{m=1}^M\) are drawn from a structurally aware distribution, but this will increase the number of computations and is outside of the scope of this paper.

\subsection{Stability}
The proposed PEs are purely generated by GNN architectures and as a result they inherit favorable stability properties of GNNs. Any stability results for GNNs hold for R-\method~as well. For instance, let $\tilde{\mathcal{G}}$ be a perturbed version of $\mathcal{G}$ such that $\tilde{\bm S}=\bm S+\bm E$. We can use the stability results in \citep{gama2020stability} and derive the following proposition. 
\begin{corollary}[Stability] \label{thm:GNNStability}
    Let $\tilde{\mathcal{G}}$ be a perturbed version of $\mathcal{G}$ such that $\tilde{\bm S}=\bm S+\bm E$ with $\left\| \bm E \right\| \leq \varepsilon$. Let $\Phi$ be an $L-$layer GNN described by Eq. (\ref{eq:GNNrec02}), where each layer consists of $F^2$ Lipschitz continuous filters [cf. Eq. (\ref{def:lipschitzFilters})] with constant $C$. Under assumptions \ref{ass:nonlinearity} and \ref{ass:filter} with $C_{\sigma}=1$ and $\beta=1/F$, it holds that:
    % eqn:genericStabilityConstant
    \begin{equation} \label{eqn:genericStabilityConstant}
     \left\|  \frac{1}{M}\sum_{m=1}^M\Phi\left( \mathcal{G},\cdot \right)\left[:,f\right]-\frac{1}{M}\sum_{m=1}^M\Phi\left( \tilde{\mathcal{G}},\cdot \right)\left[:,f\right] \right\| _{\mathcal{P}} \leq \left(1+8\sqrt{N}\right) L \varepsilon + \mathcal{O}(\varepsilon^{2})
    \end{equation}
    where $\left\|  \cdot \right\|_{\mathcal{P}}$ is the distance modulo permutation [cf. \ref{def_operator_distance}], {and $M$ is the number of samples}.
\end{corollary}
We can further normalize the proposed PEs by \(\sqrt{N} \cdot L\) to improve the stability bound. Notably, our result remains independent of the eigengap \(\delta_{\lambda}\), which is the difference between consecutive eigenvalues of the GSO. However, this independence does not hold for the stability of eigenvectors. According to the Davis-Kahan Theorem \citep{davis1970rotation}, even a small perturbation in the graph can lead to arbitrarily large differences between the eigenvector encodings of the original and perturbed GSOs. This limitation also applies to the eigenvector-based PEs in \citep{limsign}. The stability bound of the PEs in \citep{huangstability} is inversely proportional to the eigengap \(\delta_{\lambda}\) between the \(d\)-th and \((d+1)\)-th smallest eigenvalues when using the first \(d\) eigenvectors. This dependence is mitigated if all eigenvectors are computed, but doing so requires \(\mathcal{O}(N^3)\) complexity, which is impractical for large graphs. Further details on stability results, are provided in Appendix \ref{app:stability}.

\subsection{Computational complexity}
To implement R-\method, we process each initial random attribute independently using a message-passing GNN. Consequently, the computational complexity of the feed-forward pass is equivalent to that of a message-passing GNN multiplied by the number of samples, i.e., ${\Theta}\left(M NF^2+M |\mathcal{E}|F\right)$, where $F$ represents the hidden dimension of each GNN layer. The memory complexity of a serial implementation is ${\Theta}\left(NF\right)$, while for a parallel implementation, it becomes ${\Theta}\left(MNF\right)$.

\section{Our work: Basis Positional Encoding Networks (B-\method)}
The previous approach R-\method~is particularly advantageous for large graphs, where the number of samples is much smaller than the number of nodes and edges, making the computational and memory complexity approximately linear. However, for smaller-scale graphs, such as molecular graphs, the computational complexity becomes quadratic. In these cases, we propose using {standard basis vectors \(\left\{\bm{e}_m\right\}_{m=1}^{N}\) as the initial node attributes, where $\bm{e}_m [m]=1$ and $\bm{e}_m [i\neq m]=0$}, thus setting \(M = N\). Similar to the previous approach, when these samples are processed by a GNN \(\Phi\), the result is \(N\) \(\left(N \times d_p\right)\)-dimensional outputs. To maintain permutation equivariance, any equivariant function \(\rho\) can be applied, but in this paper, we choose the summation pooling for \(\rho\). Overall, B-\method~is cast as:
\begin{equation}\label{eq:BPE}
        \bm P = \rho\left[ \Phi\left( \mathcal{G}, \bm{e}_{1} \right),\dots,\Phi\left( \mathcal{G}, \bm{e}_{N} \right) \right]=\sum_{m=1}^N\Phi\left( \mathcal{G},\bm{e}_{m} \right)=\sum_{m=1}^N\bm P^{(m)}
\end{equation}
\vspace{-0.2cm}
{\begin{remark}[Expressivity, Stability]
B-\method~admits the same expressivity and stability properies as R-\method, i.e., Corollaries \ref{prop:exprpower}, and \ref{thm:GNNStability} also apply, and B-\method can count the number of 3-, 4-, 5-, 6-, and 7-node cycles in which each node participates, for any given graph. 
\end{remark}}
\vspace{-0.2cm}
\subsection{Relation to Eigenvector based encodings}
The B-\method~framework is highly related to the SPE encodings \citep{huangstability}, defined as $\text{SPE}\left(\bm V, \bm\Lambda\right)=\rho\left(\left[\bm V\text{diag}\left(\alpha_1\left(\bm\Lambda\right)\right)\bm V^T,\dots,\bm V\text{diag}\left(\alpha_F\left(\bm\Lambda\right)\right)\bm V^T\right]\right),$ where $\left\{\alpha_i\right\}_{i=1}^F$ are {continuous} functions and $\rho$ is an equivariant function. The suggested SPE implementation is $\text{SPE}\left(\bm V, \bm\Lambda\right)=\sum_{n=1}^{N}\Phi\left(\left[\bm V\text{diag}\left(\alpha_1\left(\bm\Lambda\right)\right)\bm V[n],\dots,\bm V\text{diag}\left(\alpha_M\left(\bm\Lambda\right)\right)\bm V[n]\right]\right)$,
where $\Phi$ is a GNN, and $\left\{\alpha_i\right\}_{i=1}^F$ are MLPs. The computational and memory complexity of SPE is $\mathcal{O}\left(N^3\right)$, and $\mathcal{O}\left(N^2\right)$ respectively.
\vspace{-0.1cm}
\begin{remark}
    When $\left\{\alpha_i\right\}_{i=1}^F$ in $\text{SPE}\left(\bm V, \bm\Lambda\right)$ are pointwise analytic functions, the SPE architecture is equivalent to the proposed B-\method~architecture in Eq. (\ref{eq:BPE}). The proof can be found in Appendix \ref{app:SPE}.
\end{remark}
\subsection{Computational complexity}
To implement B-\method, we process each initial basis encoding independently using a message-passing GNN. As a result, the computational complexity is \({\Theta}\left( N^2F^2 + N |\mathcal{E}| F \right)\), where \(F\) represents the hidden dimension of each GNN layer. The memory complexity for a serial implementation is \({\Theta}\left( N F \right)\), while for a parallel implementation, it increases to \({\Theta}\left( N^2 F \right)\).

\section{Experiments}
In this section, we assess the performance of \method~on graph classification and regression tasks. All experiments were conducted on a Linux server with NVIDIA A100 GPU. 
\subsection{Architectures}
To generate the proposed PE, $\Phi$ is a $9-$layer message-passing GNN with batch normalization layers and skip connections. The first layer of $\Phi$ is a generalized GNN layer, as described in Eq. (\ref{eq:GNNrec02}), and \(K\) can be greater than two. All the remaining layers in $\Phi$ are GIN layers \citep{xu2018powerful}. When $K = 2$ we omit this generalized GNN layer, and solely use GIN layers. We denote as R-\method~the architecture with random samples, described in Eq. (\ref{eq:randomPE}), and B-\method~the architecture with basis vectors, described in Eq. (\ref{eq:BPE}).
In all experiments we evaluated our model on selected values of $K$ ranging from 2 to 18, as well as different sample sizes ranging from 10 to 200, and selected the best model accordingly. The R-\method~and B-\method~encodings are fed to a \text{GINE} \citep{hu2020strategies} architecture, which is a message-passing GNN that processes node and edge attributes, as well as the graph structure and PEs. More architectural and experimental details can be found in Appendices \ref{appendix-implementation} and \ref{appendix-additional-exps}.
\subsection{Baselines}
The baseline models for comparison are grouped into four categories:  
i) \textbf{GNNs without PEs}: GCN \citep{kipf2016semi}, GIN \citep{xu2018powerful};  
ii) \textbf{GNNs with unique identifiers}: GIN with random IDs \citep{xu2018powerful, abboud2020surprising, sato2019approximation};  
iii) \textbf{GNNs with structural PEs}: GSN with cycles, GSN with cliques \citep{bouritsas2022improving};  
iv) \textbf{GNNs with eigenvector-based PEs}: SignNet, BasisNet \citep{limsign}, PEG \citep{wang2022equivariantstablepositionalencoding}, SPE \citep{huangstability}.

In addition, we implement \text{SignNet-8S}, \text{BasisNet-8S} and \text{SPE-8S} which are variants of the full SignNet, BasisNet, and SPE models. These variants employ the eigenvectors corresponding to the 8 smallest eigenvalues of the normalized Laplacian that still need $\mathcal{O}(N^3)$ computational complexity. In SPE-8S and BasisNet-8S the memory complexity remains  $\mathcal{O}(N^2)$, but in \text{SignNet-8S} it reduces from $\mathcal{O}(N^2)$ to $\mathcal{O}(N)$. Furthermore, we implement \text{SignNet-8L}, utilizing only the 8 largest eigenvectors, which reduces both the memory and computational complexity to $\mathcal{O}(N)$.
 
\subsection{Graph Classification on Social Networks}
\begin{table}[t]
%\parbox{.45\linewidth}{
\caption{Graph classification accuracy on REDDIT-B and REDDIT-M (OOM stands for out-of-memory). R-\method~outperforms all light-weight baselines by at least 2.5\% in REDDIT-B and 3.5 \% in REDDIT-M. It also achieves better performance compared to SPE in RREDDIT-M and comparable in in REDDIT-B, but with much lower complexity.
}

\label{tab:RGNN-REDDIT}
\centering
  \scalebox{0.7}{\begin{tabular}{lllcc}
    \toprule
\textbf{Method} & \textbf{Computational Complexity} & \textbf{Memory Complexity}& \textbf{REDDIT-B}& \textbf{REDDIT-M} \\
\midrule
\text{GCN}  & $\mathcal{O}\left(N\right)$ & $\mathcal{O}\left(N\right)$& $50.0 \pm 0.0$& $20.0 \pm 0.0$ \\
\text{GIN}  & $\mathcal{O}\left(N\right)$ & $\mathcal{O}\left(N\right)$& $91.8 \pm 1.0$& $56.9 \pm 2.0$ \\
\midrule
\text{GIN + rand id} & $\mathcal{O}\left(N\right)$ & $\mathcal{O}\left(N\right)$& $91.8 \pm 1.6$ & $57.0 \pm 2.1$ \\
\text{GSN with cliques} & $\mathcal{O}\left(N^2\right)$ & $\mathcal{O}\left(N\right)$&$91.1 \pm 1.8$& $56.2 \pm 1.8$ \\
\midrule
\text{SignNet-8S} & $\mathcal{O}\left(N^3\right)$ & $\mathcal{O}\left(N\right)$& $\mathbf{94.5 \pm 0.3}$& $59.3 \pm 0.5$ \\
\text{SignNet-8L} & $\mathcal{O}\left(N\right)$ & $\mathcal{O}\left(N\right)$&$ 89.0 \pm 5.2 $& $55.2 \pm 2.9$   \\
\text{SignNet-full} &$\mathcal{O}\left(N^3\right)$ & $\mathcal{O}\left(N^2\right)$& OOM & OOM   \\
\text{BasisNet} &$\mathcal{O}\left(N^3\right)$ & $\mathcal{O}\left(N^2\right)$& OOM & OOM   \\
\text{SPE} &$\mathcal{O}\left(N^3\right)$ & $\mathcal{O}\left(N^2\right)$& OOM & OOM   \\
\midrule
\text{R-\method (ours)} & $\mathcal{O}\left(MN\right)$ & $\mathcal{O}\left(N\right)/\mathcal{O}\left(MN\right)$ & $\mathbf{94.5 \pm 0.4}$& $\mathbf{60.6 \pm 0.3}$   \\
\bottomrule
  \end{tabular}}

\end{table}
We first evaluate our architecture on graph classification tasks using the REDDIT-B (2,000 graphs, 2 classes, 429.6 average nodes) and REDDIT-M (5,000 graphs, 5 classes, 508.5 average nodes) datasets \citep{yanardag2015deep}. Each graph represents an online discussion thread, with nodes representing different users, and edges indicating whether one user responded to another’s comment. In both datasets, the task is to predict the subreddit to which a particular discussion graph belongs. To train the GNN models, we conduct 10-fold cross-validation. Table \ref{tab:RGNN-REDDIT} summarizes the mean and standard deviation of classification accuracy over the 10 folds. We report the best performance observed during 350 epochs of training, as is the standard practice for this dataset. The results are presented in Table \ref{tab:RGNN-REDDIT}.

We observe that R-\method~outperforms all baselines on REDDIT-M and achieves the best performance on REDDIT-B, alongside \text{SignNet-8S}, but with one to two orders of magnitude less computational complexity. Notably, \text{SignNet-full}, \text{BasisNet}, and \text{SPE} are unable to handle these datasets due to their quadratic memory complexity.
\subsection{Graph regression on Molecular Graphs}
\begin{table}[t]
\caption{logP Prediction in ZINC. B-\method~ouperforms all the baselines both in MAE and Generalization gap. It is notable that B-\method~achieves these results with quadratic complexity compared to the second best (SPE) that operates with cubic complexity.
}
\label{tab:stability_expressiveness}
\centering
  \scalebox{0.75}{\begin{tabular}{lcccccc}
    \toprule
    \textbf{PE Method} &  \textbf{\#PEs} & \textbf{Test MAE} & \textbf{Training MAE} & \textbf{General. Gap} \\
    \midrule
    No PE  & N/A & $0.1772 \pm 0.0040$ & $0.1509 \pm 0.0086$ & $0.0263 \pm 0.0113$ \\
    \text{GCN}   & N/A & $0.469 \pm 0.002$ & $-$ &$-$ &\\
\text{GIN}  & N/A & $0.209 \pm 0.018$  & $-$ &$-$\\
\text{GSN with cycles} & 10  & $0.115 \pm 0.012$ & $-$ &$-$ \\
\text{GIN + rand id}  & 1  & $0.279 \pm 0.023$& $-$ &$-$ \\

\midrule
    
    SignNet-8S   & 8 & $0.1034 \pm 0.0056$ & $0.0418 \pm 0.0101$ & $0.0602 \pm 0.0112$ \\
    SignNet   & Full & $0.0853 \pm 0.0026$ & $0.0349 \pm 0.0078$ & $0.0502 \pm 0.0103$ \\
    BasisNet-8S   & 8 & $0.1554 \pm 0.0048$ & $0.0513 \pm 0.0053$ & $0.1042 \pm 0.0063$ \\
    BasisNet   & Full & $0.1555 \pm 0.0124$ & $0.0684 \pm 0.0202$ & $0.0989 \pm 0.0258$ \\
    SPE-8S   & 8 & $0.0736 \pm 0.0007$ & \textcolor{blue}{$0.0324 \pm 0.0058$} & $0.0413 \pm 0.0057$ \\
    SPE   & Full & \textcolor{blue}{$0.0693 \pm 0.0040$} & $0.0334 \pm 0.0054$ & $0.0359 \pm 0.0087$ \\
    \midrule
    R-\method (ours)& N/A & $0.0699 \pm 0.002$ & $0.0366 \pm 0.006$ & \textcolor{blue}{$0.0333 \pm 0.007$} \\
    B-\method (ours)   & N/A & \textcolor{red}{$0.0644 \pm 0.001$} & \textcolor{red}{$0.0290 \pm 0.003$} & \textcolor{red}{$0.0353 \pm 0.002$} \\
    \bottomrule
  \end{tabular}}
\end{table}
We also evaluate our model on the task of predicting the penalized water-octanol partition coefficient (logP) for molecules from the ZINC dataset \citep{irwin2012zinc,dwivedi2023benchmarking}. We use the standard split for this dataset, which entails 10,000 molecules for training, 1,000 for validation, and another 1,000 for testing. We report the mean and standard deviation of the MAE for the model achieving the highest validation accuracy, averaged over 4 different seeds. 
The results can be found in Table \ref{tab:stability_expressiveness}. We observe that B-\method~achieves the best results, and also the best generalization gap between the competing methods. It is also notable that R-\method~and B-\method~also outperform all the remaining competing methods.

We conduct experiments on the DrugOOD dataset \citep{ji2022drugoodoutofdistributionooddataset}. The dataset evaluates models on out-of-distribution (OOD) generalization, focusing on three specific types of domain shifts: Assay, Scaffold, and Size. The Assay, Scaffold, and Size splits test the ability to generalize to different bioassays, molecules with different structures, and molecules of different sizes respectively. The results are presented in Table \ref{tab:drugood-results}. We observe that stable methods work generally better than SignNet and BasisNet, which have reduced stability. B-\method~improves the AUC by $9.5\%$ compared to BasisNet, and $3.8\%$ compared to SignNet in Size OOD generalization. B-\method~achieves a $4.6\%$ improvement in AUC over both SignNet and BasisNet in Scaffold OOD generalization.
Furthermore, B-\method~demonstrates an advantage over SPE in size OOD generalization. This improvement is likely attributed to the linear scaling of SPE's stability bound with graph size, compared to the square-root scaling of \method's stability bound, which enhances its size generalization capabilities.

\begin{table}[h]
\centering
\caption{Binding Affinity AUROC results (5 random seeds) on DrugOOD: The performance of \method~outperforms SignNet and BasisNet and works comparably to SPE, while maintaining lower computational complexity.}
\label{tab:drugood-results}
\scalebox{0.7}{\begin{tabular}{llcccc}
    \toprule
    \textbf{Domain} & \textbf{PE Method} & \textbf{ID-Val (AUC)} & \textbf{ID-Test (AUC)} & \textbf{OOD-Val (AUC)} & \textbf{OOD-Test (AUC)} \\
    \midrule
    \multirow{6}{*}{\textbf{Assay}} 
    & No PE   & 92.92\(\pm\)0.14 & 92.89\(\pm\)0.14 & 71.02\(\pm\)0.79 & 71.68\(\pm\)1.10 \\
    & PEG     & 92.51\(\pm\)0.17 & 92.57\(\pm\)0.22 & 70.86\(\pm\)0.44 & 71.98\(\pm\)0.65 \\
    & SignNet & 92.26\(\pm\)0.21 & 92.43\(\pm\)0.27 & 70.16\(\pm\)0.56 & {\textbf{72.27\(\pm\)0.97}} \\
    & BasisNet & 88.96\(\pm\)1.35 & 89.42\(\pm\)1.18 & 71.19\(\pm\)0.72 & 71.66\(\pm\)0.05 \\
    & SPE     & 92.84\(\pm\)0.20 & 92.94\(\pm\)0.15 & 71.26\(\pm\)0.62 & \red{\textbf{72.53\(\pm\)0.66}} \\
    & SPE-8S     & 92.36\(\pm\)0.18 & 92.62\(\pm\)0.10 & 70.71\(\pm\)0.47 & 71.72 \(\pm\) 0.71 \\
    & R-\method (ours)     & 92.71\(\pm\)0.10 & 92.92\(\pm\)0.12 & 70.57\(\pm\)0.72 & \textbf{72.24\(\pm\)0.30} \\
    & B-\method (ours)     & 90.54\(\pm\)0.89 & 90.81 \(\pm\)0.79 & 70.53\(\pm\)0.67 & \text{71.22\(\pm\)0.42} \\    
    \midrule
    \multirow{6}{*}{\textbf{Scaffold}} 
    & No PE   & 96.56\(\pm\)0.10 & 87.95\(\pm\)0.20 & 79.07\(\pm\)0.97 & 68.00\(\pm\)0.60 \\
    & PEG     & 95.65\(\pm\)0.29 & 86.20\(\pm\)0.14 & 79.17\(\pm\)0.97 & \text{69.15\(\pm\)0.75} \\
    & SignNet & 95.48\(\pm\)0.34 & 86.73\(\pm\)0.56 & 77.81\(\pm\)0.70 & 66.43\(\pm\)1.06 \\
    & BasisNet & 85.80\(\pm\)3.75 & 78.44\(\pm\)2.45 & 73.36\(\pm\)1.44 & 66.32\(\pm\)5.68 \\
    & SPE     & 96.32\(\pm\)0.28 & 88.12\(\pm\)0.41 & 80.03\(\pm\)0.58 & \red{\textbf{69.64\(\pm\)0.49}} \\
    & SPE-8S     & 96.44\(\pm\)0.079 & 87.88\(\pm\)0.45 & 79.34\(\pm\)0.50 & 68.72\(\pm\)0.63 \\
    & R-\method (ours)     & 96.09\(\pm\)0.32 & 88.01\(\pm\)0.43 & 78.72\(\pm\)0.02 & \text{69.20\(\pm\)1.00} \\
    & B-\method (ours)     & 96.06\(\pm\)0.29 & 87.56\(\pm\)0.81 & 79.86\(\pm\)0.58 & {\textbf{69.51\(\pm\)0.62}} \\
    
    \midrule
    \multirow{6}{*}{\textbf{Size}} 
    & No PE   & 93.78\(\pm\)0.12 & 93.60\(\pm\)0.27 & 82.76\(\pm\)0.04 & {\textbf{66.04\(\pm\)0.70}} \\
    & PEG     & 92.46\(\pm\)0.35 & 92.67\(\pm\)0.23 & 82.12\(\pm\)0.49 & \text{66.01\(\pm\)0.10} \\
    & SignNet & 93.30\(\pm\)0.43 & 93.20\(\pm\)0.39 & 80.67\(\pm\)0.50 & 64.03\(\pm\)0.70 \\
    & BasisNet & 86.04\(\pm\)4.01 & 85.51\(\pm\)4.04 & 75.97\(\pm\)1.71 & 60.79\(\pm\)3.19 \\
    & SPE     & 92.46\(\pm\)0.35 & 92.67\(\pm\)0.23 & 82.12\(\pm\)0.49 & \text{66.02 \(\pm\) 0.49} \\
    & SPE-8S     & 93.68\(\pm\)0.20 & 93.86\(\pm\)0.12 & 83.04\(\pm\)0.63 & 65.74 \(\pm\) 2.2 \\
      & R-\method (ours)     & 93.32\(\pm\)0.34 & 93.92\(\pm\)0.20 & 82.09\(\pm\)0.44 & \text{65.89 \(\pm\) 1.30} \\
      & B-\method (ours)     &  93.18 \(\pm\) 0.45 & 93.29 \(\pm\) 0.46 & 83.14 \(\pm\) 0.37 & \red{\textbf{66.58 \(\pm\) 0.67}} \\
    \bottomrule
\end{tabular}}
\end{table}

{\subsection{Large-scale Link Prediction on Relational Databases (RelBench)}
We also test the performance of the proposed \method~on large-scale link prediction for Stack Exchange Q\&A Website Database. To that end we utilize the rel-stack dataset for the relational deep learning benchmak (RelBench) \cite{feyposition,robinson2024relbench}. Rel-stack is a temporal and heterogeneous graph with approximately 38 million nodes. We consider two different tasks; i) \texttt{user-post-comment}, where we predict a list of existing posts that a user will comment in the next two years, and ii) \texttt{post-post-related}, where we predict a list of existing posts that users will link a given post to in the next two years. The results for the two tasks can be found in Table \ref{tab:relbench_results}.}
\begin{table}[t]
{\caption{{Validation and test mean average precision (MAP) on large-scale RelBench recommendation tasks. \method~has an 11\% benefit over the backbone model with no PE on the \texttt{user-post-comment} task and a 2\% benefit on the \texttt{post-post-related} task. }}
\label{tab:relbench_results}
\centering
\scalebox{0.7}{\begin{tabular}{l p{3cm} ccccc}
    \toprule
    \text{Task} &Evaluation& No PE & SignNet-8L &SignNet-8S &{B-\method (ours)}& {R-\method (ours)} \\
    \midrule
    \multirow{2}{*}{\texttt{User-post-comment}} 
    &\raggedright\text{Val. MAP} & $15.20$ & $15.33$ & $\mathbf{15.47}$ & $15.13$ &  $15.24$ \\
    &\raggedright\text{Test MAP}&$12.47$ & $13.76$ & $13.77$& $13.80$ & $\mathbf{13.87}$ \\
    \midrule
        \multirow{2}{*}{\texttt{Post-post-related}} 
    &\raggedright\text{Val. MAP} & $8.10$ & $7.90$ &  $7.70$  & $8.00$ & $\mathbf{8.40}$ \\
    &\raggedright\text{Test MAP}&$10.73$ & $10.39$ &$10.86$&  $\mathbf{10.94}$  & $10.86$  \\
    \bottomrule
\end{tabular}}}
\end{table}

{
The backbone model for this RelBench task a heterogeneous identity-aware GNN \cite{you2021identity} and all methods are trained with batch size $20$. From Table \ref{tab:relbench_results} we observe that \method~has an 11\% benefit over the identity aware backbone model with no PE on the \texttt{user-post-comment} task and a 2\% benefit on the \texttt{post-post-related} task. \method~works similarly to SignNet-8S but with lower complexity and, and similarly to SignNet-8L on the \texttt{user-post-comment} task, but 5\% better than SignNet-8L on the \texttt{post-post-related} task.}

\section{Related work}
The works that are mostly relevant to our work can be grouped in 4 categories: i) Eigenvector-based Positional Encodings, e.g., \citep{dwivedi2020generalization,rampavsek2022recipe,kreuzer2021rethinking,mialon2021graphitencodinggraphstructure,feldman2022weisfeilerlemaninfinitespectral,huangstability,zhangexpressive}; ii) Graph Neural Networks with unique node identifiers, e.g., \citep{loukas2019graph, abboud2020surprising, sato2021random,abboud2020surprising,sato2021random,eliasof2023graphpositionalencodingrandom}; iii) Graph Representation Learning with Structural Encodings, e.g., \citep{li2020distanceencodingdesignprovably,ying2021transformers, you2019position,you2021identity,dwivedigraph,ma2023graph,kanatsouliscounting}; iv) \citep{wang2022equivariantstablepositionalencoding,srinivasanequivalence,murphy2018janossy}. A detailed discussion can be found in Appendix \ref{app:related}.

\section{Conclusion}
In this paper, we proposed a novel framework for learnable positional encodings (PEs) that addresses key limitations in existing eigenvector-based methods, particularly in terms of stability, expressive power, scalability, and genericness. By leveraging message-passing GNNs as nonlinear mappings of eigenvectors, we designed efficient PEs that maintain permutation equivariance through the use of statistical pooling functions. Our approach not only ensures high expressiveness and stability but also significantly reduces computational complexity. Experimental results demonstrate that our method consistently outperforms lightweight eigenvector-based PEs and matches the performance of full eigenvector-based methods, all while offering substantial improvements in computational efficiency. These findings open new avenues for developing scalable, expressive, and robust graph representation techniques, paving the way for advancements in graph-based learning tasks.  

\bibliography{iclr2025_conference}
\bibliographystyle{iclr2025_conference}

\appendix

\section{Related Work}\label{app:related}
% \vspace{-5cm}
\textbf{Eigenvector-based Positional Encodings:} 
Positional encodings are a crucial component in applying transformers to graph data and further integrating structural information in graph neural networks (GNNs). 
A notable approach for such positional encodings (PEs) is the use of Laplacian eigenvectors. These eigenvector-based PEs have been shown to enhance performance in transformers on graph-related tasks, as demonstrated in \citep{dwivedi2020generalization} and \citep{rampavsek2022recipe}. Additionally, they can be incorporated in attention mechanisms as seen in \citep{kreuzer2021rethinking}, \citep{mialon2021graphitencodinggraphstructure} and \citep{he2023generalization}. Laplacian eigenvectors can also be used to improve performance in the context of GNNs \citep{kim2022pure}. 

However, eigenvector-based positional encodings face challenges with stability and sign ambiguity. Small structural changes in graphs can cause significant change in eigenvectors and their corresponding positional encodings. In addition, the sign ambiguity of eigenvectors can introduce unwanted inconsistencies in these positional encodings. Works such as \cite{limsign} and \cite{huangstability} address these issues by designing sign-invariant or basis-invariant models to produce these PEs, or by making the PEs more robust and stable. \citep{zhangexpressive} introduced expressive power of spectral invariant GNNs, which are GNN architectures augmented with spectral projection matrices and provided a unified theoretical framework to analyze the previous and their proposed approach. \cite{feldman2022weisfeilerlemaninfinitespectral} used eigenvector-based heat kernels to generate node embeddings the overcome the limitations of the WL test. {\cite{geisler2024spatio} combine spatial and spectral graph filters in a unified GNN architecture.}

\textbf{Randomized Graph Neural Networks}
Initializing GNNs with unique node identifiers to enhance the expressive power has been first proposed by \citep{loukas2019graph, abboud2020surprising, sato2021random}. In particular, \citep{abboud2020surprising} and \citep{sato2021random} used random node features as inputs to GNNs, leading to enhanced function approximation, though at the expense of permutation equivariance, a key property in graph learning. \cite{eliasof2023graphpositionalencodingrandom} proposed a method for generating PEs in graph neural networks by leveraging random feature propagation, inspired by the power iteration method and its generalizations. The core of their approach involves concatenating several intermediate steps to compute the dominant eigenvectors of a propagation matrix. {\cite{duptypf} proposed a randomization method that approximates the individualization-refinement technique through particle filtering. The particle filtering GNN (PF-GNN) employs a 1-WL-based initialization method, which is subsequently refined using with particle filtering sampling to overcome the 1-WL limitations.}

\textbf{Graph Representation Learning with Structural Encodings:} Structural encodings are also important in capturing aspects of a graph's structure, such as connectivity and neighborhood information. 
\citep{li2020distanceencodingdesignprovably} uses distance PEs for GNNs (distance from an anchor node) using shortest paths and random walks. One approach is using distance-based information between nodes through methods like shortest paths or random walks, to captural structural information for transformers \citep{ying2021transformers} \citep{you2019position} \citep{you2021identity}. 
Other methods learn structural PEs directly. For instance, \citep{dwivedigraph} learn embeddings that are initialized with Laplacian eigenvectors or random walks. Similarly, \citep{ma2023graph} learn a linear combination of the Laplacian for creating relative PEs.

\textbf{Node Embedding Methods:}
One foundational approach to capturing meaningful graph representations is through node embeddings. 
DeepWalk \citep{Perozzi_2014} and node2vec \citep{grover2016node2vecscalablefeaturelearning} are early instances of these approaches and leverage random walk strategies to learn node embeddings on graphs. Although these methods show the significance of capturing structural information, they lack expressivity and do not incorporate many learnable components.

\textbf{Equivariant pooling:} Similar techniques to ours have been introduced in \citep{wang2022equivariantstablepositionalencoding}, which generate PEs by applying transformations on the Laplacian. On the other hand, \cite{kanatsouliscounting} recently analyzed the capability of GNNs to count substructures using expectation pooling functions. \cite{srinivasanequivalence} explored the equivalence between node embeddings and structural representations, showing that the expectation of node embeddings can serve as structural representations of the graph, and proposed methods to sample informative node embeddings for enhanced graph representation learning. Finally, \cite{murphy2018janossy} investigated models of permutation-invariant functions as averages of permutation-sensitive functions applied to all reorderings of a group.
% \vspace{-0.2cm}
{\subsection{Comparing \method~to Structural PEs}
The proposed \method~framework can provably count important substructures in any graph, such as cycles, cliques, and quasi-cliques. More importantly, it can generalize the counting function to graphs not seen during training, demonstrating the robust generalization ability of \method. This naturally invites comparison with methods that explicitly compute these substructures independently. Below, we summarize the key comparison points with such methods:

\textbf{Expressivity}: \method~is not limited to pre-defined motifs, such as cycles or cliques. It can compute other potentially important substructures, such as dense subgraphs, chordal cycles, or combinations of motifs, that explicit counting methods might omit simply because they are not pre-specified. Notably, the number of possible motifs in a graph grows combinatorially, highlighting the flexibility and breadth of \method.

\textbf{Complexity}: Explicitly counting high-order motifs, especially at the node level, can be computationally expensive. \method~bypasses this challenge by learning to capture these structures implicitly, making it more scalable to large and complex graphs.

\textbf{Bias}: Predetermining which motifs to count introduces bias into the model. For example, molecular graphs often benefit from detecting cycles, while social networks emphasize cliques or dense subgraphs. In contrast, \method~is task-agnostic and allows the data to guide which motifs are most relevant, adapting to the specific requirements of the application. On the flip side, when the training data have small sizes, learning can benefit by specific biases that structural PEs admit.}

\section{Proof of Proposition \ref{prop:eig}} \label{app:MLP}
Under the assumptions of Proposition \ref{prop:eig} the GNN has the following recursive formula:
\begin{align}
\bm X^{(l)} &= \texttt{MLP}\left(\bm X^{(l-1)},\bm S\bm X^{(l-1)}\bm\right)=\texttt{MLP}^{(-1)}\left(\sigma\left(\sum_{k=0}^{K-1} \bm S^k\bm X^{(l-1)}\bm H_k^{(l)}\right)\right),
\end{align}
where $\texttt{MLP}^{(-1)}$ denotes the all the layers of $\texttt{MLP}$ except the first layer, and $K=2$. We know compute the eigenvalue decomposition of $\bm S^k = \bm V\bm\Lambda^k\bm V^T$, and use some extra algebraic manipulations.
\begin{align}
\bm X^{(l)} &=\texttt{MLP}^{(-1)}\left(\sigma\left(\sum_{k=0}^{K-1} \bm V\bm \Lambda^k\bm V^T\bm X^{(l-1)}\bm H_k^{(l)}\right)\right)\\&=\texttt{MLP}^{(-1)}\left( \sigma\left(\sum_{k=0}^{K-1} \sum_{n=1}^N\lambda_n^k\bm v_n\bm v_n^T \bm X^{(l-1)}\bm H_k^{(l)}\right)\right)\\&= \texttt{MLP}^{(-1)}\left(\sigma\left(\sum_{k=0}^{K-1} \sum_{n=1}^N\lambda_n^k\bm v_n\left[\bm v_n^T \bm X^{(l-1)}\left[:,1\right],\dots,\bm v_n^T \bm X^{(l-1)}\left[:,F_{l-1}\right]\right]\bm H_k^{(l)}\right)\right),
\end{align}
where $\bm V[:,n] = \bm v_n$ and $\bm \Lambda[n,n] = \lambda_n$. We now focus on the output of the first MLP layer $\bm X^{(l,1)}$, $\bm X^{(l)}=\texttt{MLP}^{(-1)}\left(\bm X^{(l,1)}\right)$ first layer only $\bm$: Then each feature of $\bm X^{(l,1)}$ can be written as:
\begin{align}
\bm X^{(l,1)}[:,f]&= \sigma\left(\sum_{k=0}^{K-1} \sum_{n=1}^N\lambda_n^k\bm v_n\left[\bm v_n^T \bm X^{(l-1)}\left[:,1\right],\dots,\bm v_n^T \bm X^{(l-1)}\left[:,F_{l-1}\right]\right]\bm H_k^{(l)}[:,f]\right)
 \\&= \sigma\left(\sum_{k=0}^{K-1} \sum_{n=1}^N\lambda_n^k\bm v_n\sum_{i=1}^{F_{l-1}}\bm H_k^{(l)}[i,f]\bm v_n^T \bm X^{(l-1)}\left[:,i\right]\right)\\&= \sigma\left( \sum_{n=1}^N\sum_{i=1}^{F_{l-1}}\sum_{k=0}^{K-1}\lambda_n^k\bm H_k^{(l)}[i,f]<\bm v_n, \bm X^{(l-1)}\left[:,i\right]>\bm v_n\right)\\&= \sigma\left( \sum_{n=1}^N\bm W\left[n, f\right]\bm v_n\right),
\end{align}
where:
\begin{equation}
\bm W\left[n, f\right] = \sum_{i=1}^{F_{l-1}}\sum_{k=0}^{K-1}\lambda_n^k\bm H_k^{(l)}[i,f]<\bm v_n, \bm X^{(l-1)}\left[:,i\right]>.
\end{equation}
As a result $\bm X^{(l)}=\sigma\left(\bm V\bm W\right)$. When $\bm S$ is not symmetric we can replace $\bm v_n$ to $\bm V^{-1}[n,:]$. This concludes the proof.

{\section{Basis Universality of \method}\label{app:universality}
We consider the general form of \method:
\begin{equation}
        \bm P = \rho\left[ \Phi\left( \mathcal{G},\bm q^{(1)} \right),\dots,\Phi\left( \mathcal{G},\bm q^{(M)} \right) \right]\in \mathbb{R}^{N \times d_p},
\end{equation}
where $\rho$ is an unrestricted equivariant function and $\Phi$ is a message-passing GNN with skip connections. We let $\bm q^{(m)} = \bm e_m$ and $M=N$. From Proposition \ref{prop:eig} we get that:

\begin{equation}
\bm X^{(l)} =\texttt{MLP}\left(\bm V\right)=\texttt{MLP}^{(-1)}\left(\sigma\left(\bm V \bm W\right)\right)
\end{equation}
\begin{equation}
\bm W\left[n, f\right] = \sum_{i=1}^{F_{l-1}}\sum_{k=0}^{K-1}\lambda_n^k\bm H_k^{(l)}[i,f]\langle \bm v_n, \bm X^{(l-1)}\left[:,i\right]\rangle,
\end{equation}

We ommit the nonlinearities from the GNN and for $\bm X^{(0)}=\bm e_m$ we get:

\begin{equation}
\bm X^{(K)}=\bm V \bm W,\quad\bm W\left[n, f\right] = \langle \bm v_n, \bm e_m\rangle \sum_{k=0}^{K-1}\bm h_k[f]\lambda_n^k,
\end{equation}

As a result $\bm W\left[n, f\right]$ is a polynomial on the eigenvalues $\tilde{h}_f\left(\lambda_n\right)=\sum_{k=0}^{K-1}\bm h_k[f]\lambda_n^k$ scaled by $\langle \bm v_n, \bm e_m\rangle$. We will then use the following lemma.

\begin{lemma}\label{lemma:filter}
Let $\mathcal{G}$ be a graph with $N$ nodes and GSO $\bm S\in\mathbb{R}^{N\times N}$. Also let $\mathcal{S}=\left\{\left\{\lambda_1,\dots,\lambda_N\right\}\right\}$ be the multiset of eigenvalues of $\bm S$; $\mathcal{S}$ can have repeated elements (eigenvalues). Also, let $\mathcal{M}=\{\mu_1,\dots,\mu_q\}$ be the ordered set of all distinct (non-repeated) eigenvalues of $\bm S$. We can always design poynomial filter such that:
\begin{equation}\label{eq:negativefilter}
    \tilde{h}\left(\lambda\right) = \bigg\{
\begin{matrix}
\hspace{-0.8cm}\gamma\left(\mu_f\right),~\text{if}~~\lambda=\mu_f\\
~~0\quad\quad,~\text{if}~~\lambda=\mu_f\neq \mu_i
\end{matrix}
\end{equation}
Proof: Let 
\begin{align}
\begin{bmatrix}
\tilde{h}\left(\mu_1\right)\\
\tilde{h}\left(\mu_2\right)\\
\vdots\\
\tilde{h}\left(\mu_q\right)
\end{bmatrix}=\begin{bmatrix}
1~\mu_1~\mu_1^2 \dots \mu_1^{K-1}\\
1~\mu_2~\mu_2^2 \dots \mu_2^{K-1}\\
\vdots\\
1~\mu_q~\mu_q^2 \dots \mu_q^{K-1}\\
\end{bmatrix}\begin{bmatrix}
h_0\\
h_1\\
\vdots\\
h_{K-1}
\end{bmatrix} = \bm B \bm h
\end{align}
$\bm B$ is a Vandermonde matrix and when $K=q$ the determinant of $\bm B$ takes the form:
\begin{equation}
    \text{det}\left(\bm B\right) = \prod_{1\leq i<j\leq q} \left(\mu_i-\mu_j\right)
\end{equation}
Since the values $\mu_i$ are distinct, $\bm B$ has full column rank and there exists a polynomial $\tilde{h}$ with unique parameters $\bm h =\bm B^{-1}\bm e_{i} \gamma\left(\mu_f\right)$ such that $\tilde{h}\left(\lambda\right) =\gamma\left(\mu_f\right)$ if $\lambda=\mu_f$, and $\tilde{h}\left(\lambda\right) =0$ if $\lambda=\mu_j\neq \mu_f$.
\end{lemma}

Using Lemma \ref{lemma:filter}, we can design $\tilde{h}_f\left(\lambda_n\right)=\sum_{k=0}^{K-1}\bm h_k[f]\lambda_n^k$ such that:

\begin{equation}\label{eq:negativefilter2}
    \tilde{h}_f\left(\lambda_n\right) = \bigg\{
\begin{matrix}
\hspace{-0.8cm}1,~\text{if}~~\lambda_n=\mu_f\\
0,~\text{if}~~\lambda_n=\mu_j\neq \mu_f
\end{matrix}
\end{equation}
Under this parametrization, $\bm X^{(K)}$ takes the form:
\begin{equation}
    \bm X^{(K)}=\begin{bmatrix}
        \bm V_{\mu_1}\bm V_{\mu_1}^T\bm e_m,\dots, \bm V_{\mu_q}\bm V_{\mu_q}^T\bm e_m
    \end{bmatrix}\in\mathbb{R}^{N\times q},
\end{equation}
where $\bm V_{\mu_f}$ is the eigenspace (orthogonal space of the eigenvectors) corresponding to eigenvalue $\mu_f$. Since we independently feed $\bm e_1,\dots, \bm e_N$ to the \method~architecture, we will have $N$ output samples for each output feature, i.e., $N$ samples for $\bm X^{(K)}[:,f]$. We will represent the $m-$th sample as $\bm X^{(K)}[:,f,m]$ and for the $f-$th output feature will have the following samples:
\begin{equation}
    \bm X^{(K)}[:,f,:]=\begin{bmatrix}
        \bm V_{\mu_f}\bm V_{\mu_f}^T\bm e_1,\dots, \bm V_{\mu_f}\bm V_{\mu_f}^T\bm e_N
    \end{bmatrix}=\bm V_{\mu_f}\bm V_{\mu_f}^T,
\end{equation}
We process the output samples of each feature via an equivariant function $\rho$, to get the final output embedding as:
\begin{equation}
    \bm Y=\rho\left(
        \bm V_{\mu_1}\bm V_{\mu_1}^T,\dots, \bm V_{\mu_q}\bm V_{\mu_q}^T
    \right).
\end{equation}
We can choose $\rho$ to operate on each feature independently i.e.,
\begin{equation}\label{eq:Bnet}
    \bm Y=\rho\left(
        g^{(1)}\left(\bm V_{\mu_1}\bm V_{\mu_1}^T\right),\dots,g^{(q)}\left(\bm V_{\mu_q}\bm V_{\mu_q}^T\right)
    \right),
\end{equation}
where $\left\{g^{(i)}\right\}_{i=1}^f$ is universally approximating permutation
equivariant or invariant function, e.g., high-order tensor IGN \citep{maron2018invariant}.
Equation (\ref{eq:Bnet}) is the definition of BasisNet \citep{limsign}. BasisNet universally approximates all continuous basis invariant functions, which proves that \method~is also a universal approximator of basis invariant functions.}

{\section{\method~is a universal function of eigenvalues}
{\begin{theorem}[Eigenvalue Universality]
    Let $\mathcal{G}$ be a graph with GSO $\bm S = \bm V\bm\Lambda\bm V^T$, and $f$ be any continuous function of eigenvalues $f\left(\diag\left(\bm\Lambda\right)\right)$. Then there exist GNN $\Phi$ and a continuous invariant function $\rho$, such that $f\left(\diag\left(\bm\Lambda\right)\right) = \rho\left[ \Phi\left( \mathcal{G},\bm q^{(1)} \right),\dots,\Phi\left( \mathcal{G},\bm q^{(M)} \right) \right]$. 
\end{theorem}}

\emph{Proof}: Going back to Eq. \ref{eq:Bnet} we can set $\rho\circ g^{(i)}=\bm 1^T\diag\left(\cdot\right)$, which is an invariant function that performs graph pooling. Then the output $\bm y\in\mathbb{R}^q$ takes the form:
\begin{align}
    \bm y&=\left[\bm 1^T\diag\left(\bm V_{\mu_1}\bm V_{\mu_1}^T\right),\dots,\bm 1^T\diag\left(\bm V_{\mu_q}\bm V_{\mu_q}^T\right)\right]=\left[\bm 1^T\left(\bm V_{\mu_1}\right).^2\bm 1,\dots,\bm 1^T\left(\bm V_{\mu_q}\right).^2\bm 1\right]\\
    &=\left[\text{mult}\left(\mu_1\right),\dots,\text{mult}\left(\mu_q\right)\right],
\end{align}
where $\text{mult}\left(\mu_i\right)$ is the multiplicity of eigenvalue $\mu_i$.

Using Lemma \ref{lemma:filter}, we can design $\tilde{h}_f\left(\lambda_n\right)=\sum_{k=0}^{K-1}\bm h_k[f]\lambda_n^k$ such that:
\begin{equation}\label{eq:negativefilter22}
    \tilde{h}_f\left(\lambda_n\right) = \bigg\{
\begin{matrix}
\hspace{-0.5cm}\mu_f,~\text{if}~~\lambda_n=\mu_f\\
~~0,~\text{if}~~\lambda_n=\mu_j\neq \mu_f
\end{matrix}
\end{equation}
As a result we can design a set of $2q$ filters $\left\{\tilde{h}_f\right\}_{f=1}^{2q}$ such that:
\begin{align}
    \tilde{h}_{2f-1}\left(\lambda_n\right) &= \bigg\{
\begin{matrix}
\hspace{-0.5cm}\mu_f,~\text{if}~~\lambda_n=\mu_f\\
~~0,~\text{if}~~\lambda_n=\mu_j\neq \mu_f
\end{matrix}\\
    \tilde{h}_{2f}\left(\lambda_n\right) &= \bigg\{
\begin{matrix}
\hspace{-0.7cm}1,~\text{if}~~\lambda_n=\mu_f\\
~~0,~\text{if}~~\lambda_n=\mu_j\neq \mu_f
\end{matrix}
\end{align}

Under this parametrization, $\bm X^{(K)}$ takes the form:
\begin{equation}
    \bm X^{(K)}=\begin{bmatrix}
        \bm V_{\mu_1}\mu_1\bm V_{\mu_1}^T\bm e_m,\bm V_{\mu_1}\bm V_{\mu_1}^T\bm e_m,\dots,\bm V_{\mu_q}\mu_q\bm V_{\mu_q}^T\bm e_m, \bm V_{\mu_q}\bm V_{\mu_q}^T\bm e_m
    \end{bmatrix}\in\mathbb{R}^{N\times 2q},
\end{equation}
and according to previous analysis we get:
\begin{equation}
    \bm Y=\rho\left(
        \bm V_{\mu_1}\mu_1\bm V_{\mu_1}^T,\bm V_{\mu_1}\bm V_{\mu_1}^T,\dots, \bm V_{\mu_q}\mu_q\bm V_{\mu_q}^T,\bm V_{\mu_q}\bm V_{\mu_q}^T
    \right).
\end{equation}
We can choose $\rho$ to be an invariant function that involves graph pooling operations as:
\begin{equation}
    \rho=\text{MLP}\left[\bm 1^T\diag\left(\cdot\right),\dots,\bm 1^T\diag\left(\cdot\right)\right]
\end{equation}
Then the output $\bm y\in\mathbb{R}^{2q}$ takes the form: 
\begin{align}
    \bm y&=\small{\text{MLP}\left[\bm 1^T\diag\left(\bm V_{\mu_1}\mu_1\bm V_{\mu_1}^T\right),\bm 1^T\diag\left(\bm V_{\mu_1}\bm V_{\mu_1}^T\right),\dots,\bm 1^T\diag\left(\bm V_{\mu_q}\mu_q\bm V_{\mu_q}^T\right),\bm 1^T\diag\left(\bm V_{\mu_q}\bm V_{\mu_q}^T\right)\right]}\nonumber\\&=\text{MLP}\left[\bm 1^T\left(\bm V_{\mu_1}\right).^2\bm 1\mu_1,\bm 1^T\left(\bm V_{\mu_1}\right).^2\bm 1,\dots,\bm 1^T\left(\bm V_{\mu_q}\right).^2\bm 1\mu_q,\bm 1^T\left(\bm V_{\mu_q}\right).^2\bm 1\right]\\
&=\text{MLP}\left[\text{mult}\left(\mu_1\right)\mu_1,\text{mult}\left(\mu_1\right)\dots,\text{mult}\left(\mu_q\right)\mu_q,\text{mult}\left(\mu_q\right)\right],
\end{align}
which is a universal function of eigenvalues. This concludes our proof.}

\section{Vertex and frequency domain analysis of \method}\label{app:taylor}
Let the input to the GNN encoder \(\bm{q} = \left[q_1, \dots, q_N\right]\) be a set of \(N\) i.i.d. random variables such that \(\mathbb{E}[q_i] = 0\) and \(\mathbb{E}[q_i^p] = 1\) for \(p \geq 2\). As shown in Eq. (\ref{eq:GNNmulti3}) $\bm q$ is processed by a set of functions:
\begin{equation}\label{eq:graphfiltermain2}
\bm y =\sigma\left(\bm z\right)=\sigma\left(\sum_{k=0}^{K-1}h_k\bm S^k\bm q\right) =\sigma\left(\bm H\left(\bm S\right)\bm q\right).
\end{equation} 
Now we assume that the pointwise nonlinearity is analytic and expand it as a Taylor series:
\begin{equation}\label{eq:graphfiltermain22}
\bm y =\sigma\left(\bm z\right)=\sum_{n=0}^{\infty}\beta_n\bm z^{n}=\sum_{n=0}^{{\infty}}\beta_n\left(\bm H\left(\bm S\right)\bm q\right)^n,
\end{equation} 
where $\beta_n=\frac{\sigma^{(n)}\left(0\right)}{n!}$. If we only use one layer, each feature of our RPEs will be:
\begin{align}\label{eq:graphfiltermain3}
\bm p =&\mathbb{E}\left[\bm y\right] = \mathbb{E}\left[\sigma\left(\bm z\right)\right]=\sum_{n=0}^{{\infty}}\beta_n\mathbb{E}\left[\bm z^n\right]=\sum_{n=0}^{{\infty}}\beta_n\mathbb{E}\left[\left(\bm H\left(\bm S\right)\bm q\right)^n\right]\\=&
\sum_{n=0}^{{\infty}}\beta_n\underbrace{\bm H\left(\bm S\right)*\dots *\bm H\left(\bm S\right)}_{n~\text{times}}\bm 1=\sum_{n=0}^{{\infty}}\sum_{i_1,i_2,\dots,i_n}\beta_n h_{i_1}\dots h_{i_n}\left(\bm S^{i_1}*\dots *\bm S^{i_n}\right)\bm 1
\end{align} 
where $*$ represents the Hadamard product. As a result, the produced PE is a linear combination of the following features $\left(\bm S^{i_1}*\dots *\bm S^{i_n}\right)\bm 1$. Using more layers to produce the proposed PEs yields more complex functions. As we proved in Proposition \ref{prop:eig}, a GNN operates as a nonlinear function of eigenvectors. To exactly analyze the proposed PEs as functions of eigenvectors let $\bm S = \bm V\bm \Lambda\bm V^T$, be the eigendecomposition of the GSO $\bm V$. Then we can show that:
\begin{align}\label{eq:moment2}
    \mathbb{E}\left[\bm z^n\right]=\sum_{i_1,\dots,i_m=0}^{K}h_{i_1,\dots,i_n}\left({\bm V}^{i_1^T}\odot\dots\odot\bm V^{i_n^T}\right)^T\left({\bm \Lambda}^{i_1^T}\otimes\dots\otimes\bm \Lambda^{i_n^T}\right)\left({\bm V}^{i_1^T}\odot\dots\odot\bm V^{i_n^T}\right)\bm 1,
\end{align}
where $\otimes$ represents the Kronecker product and $\odot$ represents the Khatri-Rao product (columnwise Kronecker). The Equation in (\ref{eq:moment2}) is a linear combination of eigenvector ``monomials''. In other words Eq. (\ref{eq:moment2}) instantiates Hadamard products of different eigenvector combinations and linearly combines them.

\section{Sample Complexity}\label{app:samplecomplexity}

To prove Theorem \ref{prop:sample} and characterize the sample complexity of our approach we will use this version of Chebychef's inequality \citep{boucheron2003concentration}
 as:
 % \frac{1}{M}\sum_{m=1}^M\Phi\left( \mathcal{G},\bm q^{(m)} \right) - \mathbb{E}\left[\Phi\left( \mathcal{G},\bm q \right)\right]
\begin{equation}
 P\left( \frac{1}{M}\left| \sum_{m=1}^M \left( \bm P^{(m)} - \mathbb{E}[ \Phi\left( \mathcal{G},\bm q \right)]\right) \right| \geq \epsilon \right) \leq \frac{\text{var}\left(\Phi\left( \mathcal{G},\bm q\right)\right)}{M\cdot \epsilon^2}. 
\end{equation}
To establish a bound for the variance of the output \( \Phi(\mathcal{G}, \bm{q}) \), we begin by analyzing how pointwise nonlinearity affects the variance of a random variable. Let \( X \) be a random variable with variance \( \text{Var}(X) \), and let \( \sigma \) be a Lipschitz continuous function with Lipschitz constant \( C_{\sigma} \). Our goal is to examine the impact of applying \( \sigma \) to \( X \), specifically focusing on how it influences the variance of the transformed variable \( \sigma(X) \).
\subsection{Effect of pointwise activation to the variance of a random variable}
   Since \( \sigma \) is Lipschitz continuous with constant \( C_{\sigma} \), for any values of \( X \) and \( \mathbb{E}[X] \), we can apply the Lipschitz condition:
   \[
   |\sigma\left(X\right) - \sigma\left(\mathbb{E}[X]\right)| \leq C_{\sigma} |X - \mathbb{E}[X]|.
   \]
   
   Taking squares on both sides:
   \[
   \left(\sigma\left(X\right) - \sigma\left(\mathbb{E}[X]\right)\right)^2 \leq C_{\sigma}^2 \left(X - \mathbb{E}[X]\right)^2.
   \]

   Now, take the expectation of both sides:
   \[
   \mathbb{E}[\left(\sigma\left(X\right) - \sigma\left(\mathbb{E}[X]\right)\right)^2] \leq C_{\sigma}^2 \mathbb{E}[\left(X - \mathbb{E}[X]\right)^2].
   \]
   
   Since \( \mathbb{E}[\left(X - \mathbb{E}[X]\right)^2] = \text{Var}\left(X\right) \), this simplifies to:
   \[
   \mathbb{E}[\left(\sigma\left(X\right) - \sigma\left(\mathbb{E}[X]\right)\right)^2] \leq C_{\sigma}^2 \text{Var}\left(X\right).
   \]

Now let's work on the left-hand side of the previous equation:
\begin{align}
    \mathbb{E}[\left(\sigma\left(X\right) - \sigma\left(\mathbb{E}[X]\right)\right)^2] &= \mathbb{E}\left[\left(\sigma\left(X\right) - \mathbb{E}[\sigma\left(X\right)] + \mathbb{E}[\sigma\left(X\right)] - \sigma\left(\mathbb{E}[X]\right)\right)^2\right]\\ &= \mathbb{E}[\left(\sigma\left(X\right) - \mathbb{E}\left[\sigma\left(X\right)\right]\right)^2] + \left(\mathbb{E}[\sigma\left(X\right)] - \sigma\left(\mathbb{E}[X]\right)\right)^2\\ &+2\mathbb{E}[<\left(\sigma\left(X\right) - \mathbb{E}[\sigma\left(X\right)]\right), \left(\mathbb{E}\left[\sigma\left(X\right)] - \sigma\left(\mathbb{E}[X]\right)\right)>\right]\\ &= \mathbb{E}[\left(\sigma\left(X\right) - \mathbb{E}\left[\sigma\left(X\right)\right]\right)^2] + \left(\mathbb{E}[\sigma\left(X\right)] - \sigma\left(\mathbb{E}[X]\right)\right)^2\\ &= \text{Var}\left(\sigma\left(X\right)\right) +  \left(\mathbb{E}[\sigma\left(X\right)] - \sigma\left(\mathbb{E}[X]\right)\right)^2,
\end{align}

Now let $\mu = \left(\mathbb{E}[\sigma\left(X\right)] - \sigma\left(\mathbb{E}[X]\right)\right)$, then the variance of \( \sigma(X) \) is bounded by:
\begin{equation}\label{eq:sc_proof1}
    \text{Var}(\sigma(X)) \leq C_{\sigma}^2 \text{Var}(X)-\mu^2\leq C_{\sigma}^2 \text{Var}(X).
\end{equation}

This shows that the Lipschitz constant \( C_{\sigma} \) acts as a scaling factor on the variance of the random variable. If \( C_{\sigma} \) is large, the variance of \( \sigma(X) \) can be significantly larger, and if \( C_{\sigma} \) is small, it can shrink the variance accordingly. For the majority of nonlinearities used in deep learning, as ReLU, sigmoid, and hyperbolic tangent, $ C_{\sigma}=1$, and $\text{Var}(\sigma(X)) \leq \text{Var}(X).$
\subsection{Effect of Graph Convolution to the variance of a random node signal}
The next step is to study the effect of graph convolution (linear message-passing) operations to a set of node features. In particular, let $\bm X^{(l)}\in\mathbb{R}^{N\times F_{l-1}}$ be the node input to the $l-th$ GNN layer. Then we define $\bm Z^{(l)}\in\mathbb{R}^{N\times F_l}$ as:
\begin{equation}\label{eq:GNNmulti}
    \bm X^{(l)} =  \sigma\left(\bm Z^{(l)}\right),~\bm Z^{(l)} = \sum_{k=0}^{K-1}\bm S^k \bm X^{(l-1)}\bm H_k
\end{equation}
After some algebraic manipulations, we can see that:
\begin{equation}\label{eq:GNNmulti2}
    \bm Z^{(l)} = \sum_{k=0}^{K-1}\bm S^k \sum_{f=1}^{F_{l-1}}\bm X^{(l-1)}[:,f]\bm H_k[f,:]^T= \sum_{f=1}^{F_{l-1}}\sum_{k=0}^{K-1}\bm S^k \bm X^{(l-1)}[:,f]\bm H_k[f,:]^T,
\end{equation}
and each feature of $\bm Z^{(l)}$ can be cast as:
\begin{equation}\label{eq:GNNmulti3}
    \bm Z^{(l)}[:,d] = \sum_{f=1}^{F_{l-1}}\sum_{k=0}^{K-1}\bm H_k[f,d]\bm S^k \bm X^{(l-1)}[:,f],~~d\in\{1,\dots,F_l\}.
\end{equation}
The above equation implies that each feature $\bm Z^{(l)}[:,d]$ is generated by a summation over $F_{l-1}$ features of type:
\begin{equation}
    \bm z = \sum_{k=0}^{K-1}\bm h_k\bm S^k \bm x=\bm H\left(\bm S\right)\bm x
\end{equation}
We assume that norm of $\bm H\left(\bm S\right)=\sum_{k}h_k\bm S^k$ is bounded, i.e., $\lVert\bm H\left(\bm S\right)\rVert\leq \beta$.

As a result, we will first  analyze the variance of $\bm z$ when the input $\bm x$ has covariance matrix:
\begin{equation}
\mathbb{E}\left[\left(\bm x-\mathbb{E}\left[\bm x\right]\right)\left(\bm x-\mathbb{E}\left[\bm x\right]\right)^T\right] = \bm C
\end{equation}
The covariance of $\bm z$ is written as:
\begin{align}
\mathbb{E}\left[\left(\bm z-\mathbb{E}\left[\bm z\right]\right)\left(\bm z-\mathbb{E}\left[\bm z\right]\right)^T\right] &= \bm H\left(\bm S\right)\bm C\bm H\left(\bm S\right)=\sum_{k=0}^{K-1}h_k\bm S^k\bm C\sum_{m=0}^{K-1}h_m\bm S^m\\&=\sum_{k=0}^{K-1}\sum_{m=0}^{K-1}h_k h_m\bm S^k\bm C\bm S^m
\end{align}
and the variance for each individual variable $\bm z[i]$ is:
\begin{align}
    \text{var}\left(\bm z\left[i\right]\right) &=\sum_{k=0}^{K-1}\sum_{l=0}^{K-1}h_k h_l\bm S^k[i,:]^T\bm Q\bm S^l[:,i]\\&=\sum_{k=0}^{K-1}\sum_{l=0}^{K-1}h_k h_l\sum_{m\in\mathcal{N}_i^{(k)}}\sum_{n\in\mathcal{N}_i^{(l)}}\bm S^k[i,m]\bm S^l[i,n]\text{cov}\left(\bm x\left[m\right],\bm x\left[n\right]\right)
    % \end{align}
% {We now consider two different scenarios. The first assumes the $\bm S$ is a GSO with non-negative, e.g., adjacency, normalized adjacency, random walk matrix.}
%   \begin{align}  
%     \text{var}\left(\bm z\left[i\right]\right) 
    \\&\leq\sum_{k=0}^{K-1}\sum_{l=0}^{K-1}h_k h_l\sum_{m\in\mathcal{N}_i^{(k)}}\sum_{n\in\mathcal{N}_i^{(l)}}\bm S^k[i,m]\bm S^l[i,n]\left|\text{cov}\left(\bm x\left[m\right],\bm x\left[n\right]\right)\right|\\&\leq\sum_{k=0}^{K-1}\sum_{l=0}^{K-1}h_k h_l\sum_{m\in\mathcal{N}_i^{(k)}}\sum_{n\in\mathcal{N}_i^{(l)}}\bm S^k[i,m]\bm S^l[i,n]\max_i\left(\text{var}\left(\bm x\left[i\right]\right)\right)\label{in:CS}\\&\leq\sum_{k=0}^{K-1}\sum_{l=0}^{K-1}h_k h_l\text{deg}^k_{\max}\text{deg}^l_{\max}\max_i\left(\text{var}\left(\bm x\left[i\right]\right)\right)\label{in:power}\\&\leq\sum_{k=0}^{K-1}h_k\text{deg}^k_{\max}\sum_{l=0}^{K-1} h_l\text{deg}^l_{\max}\max_i\left(\text{var}\left(\bm x\left[i\right]\right)\right)\leq \beta^2\max_i\left(\text{var}\left(\bm x\left[i\right]\right)\right)\label{in:bound},
\end{align}
{where $\text{deg}^k_{\max}$ is the maximum degree of $\bm S^{k}$, and is equal to $\text{deg}^k_{\max}=1$, when $\bm S$ is the normalized adjacency matrix or the random walk matrix.} The inequality in (\ref{in:CS}) comes from the Cauchy-Schwartz inequality,  the inequality  in (\ref{in:power}) comes from the definition of $\bm S^k$, and the last inquality in (\ref{in:bound}) is due the boundedness of the operator $\bm H\left(S\right)$.

Overall,
\begin{equation}\label{eq:sc_proof2}
    \text{var}\left(\bm z\left[i\right]\right) \leq \beta^2\max_i\left(\text{var}\left(\bm x\left[i\right]\right)\right).\end{equation}

The final step is to analyze the the variance of a random variable that is a sum of dependent random variables, $\bm z[i] = \sum_{f=1}^{F_{l-1}} \bm z_f[i]$. Then:
\begin{align}\label{eq:sc_proof3}
    \text{var}\left(\bm z\left[i\right]\right) &=\mathbb{E}\left[\left(\sum_{f=1}^{F_{l-1}} \bm z_f[i]\right)^2\right]=\sum_{f=1}^{F_{l-1}} \sum_{g=1}^{F_{l-1}} \mathbb{E}\left[\bm z_f[i],\bm z_g[i]\right]\leq \sum_{f=1}^{F_{l-1}} \sum_{g=1}^{F_{l-1}} \left|\mathbb{E}\left[\bm z_f[i],\bm z_g[i]\right]\right|\\&\leq F_{l-1}^2\max_f\left(\text{var}\left(\bm z_f\left[i\right]\right)\right),
\end{align}
where the last inequality in (\ref{in:CS}) comes from the Cauchy-Schwartz inequality. which is quadratic with respect to the length of the GNN layer. Combining Eq. (\ref{eq:sc_proof1}), (\ref{eq:sc_proof2}),~and (\ref{eq:sc_proof3} we conclude that:
\begin{equation}
    \text{var}\left[\bm X^{(l)}\right] \leq C^2_{\sigma}\beta^2F_{l-1}^2\max\left(\text{var}\left[\bm X^{(l-1)}\right]\right),
\end{equation}
If we assume the $F_l=F$ for all hidden layers, we get that:
 
\begin{equation}
    \text{var}\left[\bm X^{(L)}\right] \leq \left(C_{\sigma}\beta F\right)^{2L}\max\left(\text{var}\left[\bm X\right]\right),
\end{equation}
In our proposed approach $\max\left(\text{var}\left[\bm X\right]\right)=1$. If we further assume that $C_{\sigma}=1$, which is usually the case in practice, and that $\beta = 1/F$, which means that the magnitude of trainable parameters is inversely proportional to the number of hidden dimensions in each layer, we get that: \begin{equation}
    \text{var}\left[\bm X^{(L)}\right] \leq 1.
\end{equation}
This concludes the proof for Theorem \ref{prop:sample}, which is repeated below:
\begin{theorem}[Sample Complexity]
Let $\bm P$ denote the output of the architecture described in Eq. (\ref{eq:randomPE}), for a graph $\mathcal{G}$ with i.i.d. initial node attributes with unit variance. Also let $\Phi$ be an $L-$layer GNN described by Eq. (\ref{eq:GNNrec02}), with $F$ hidden dimensions at each layer. If $C_{\sigma}=1$ and $\beta=1/F$, the number of samples required such that:
\begin{equation}
    \left|  \frac{1}{M}\sum_{m=1}^M\Phi\left( \mathcal{G},\bm q^{(m)} \right) - \mathbb{E}\left[\Phi\left( \mathcal{G},\bm q \right)\right] \right| < \epsilon,~\text{with probability at least}~1-\delta,
\end{equation}
satisfies:
\begin{equation}
M \leq \frac{1}{\delta\cdot \epsilon^2}.
\end{equation}
\end{theorem}

{Theorem \ref{prop:sample} bounds the number of samples required for the output of R-\method~to approximate the expected value $\mathbb{E}\left[\Phi(\mathcal{G}, \mathbf{q})\right]$, where $\mathbf{q}\in\mathbb{R}^N$ is a random vector. It also describes the number of samples requqired for R-\method to preserve equivariance. To be more precise, since $\mathbf{q}$ is processed using an equivariant GNN $\Phi(\mathcal{G}, \cdot)$, the resulting distribution of $\Phi(\mathcal{G}, \mathbf{q})$ is permutation equivariant. Consequently, the expected value $\mathbb{E}\left[\Phi(\mathcal{G}, \mathbf{q})\right]$ is also permutation equivariant.

R-PEARL computes the empirical mean $\hat{\mathbb{E}}\left[\Phi(\mathcal{G}, \mathbf{q})\right]$, enabling it to approximate $\mathbb{E}\left[\Phi(\mathcal{G}, \mathbf{q})\right]$ with high precision. In essence, R-PEARL achieves equivariance with increasing accuracy as the number of samples $M$ grows. For very small sample sizes, R-PEARL is not equivariant; however, in practice, a modest number of samples is sufficient to render it effectively equivariant. This is evident in our experiments. In Tables 1 and 2, the "GIN + rand id" model corresponds to R-PEARL with a single sample, which is not equivariant. We observe that "GIN + rand id" performs significantly worse—more than four times worse in logP prediction for molecular graphs—compared to R-PEARL.}

\section{Proof of Corollary \ref{thrm:cycles7}} \label{app:expr}
To prove Corollary \ref{thrm:cycles7}, we assume that the pointwise nonlinearities \(\sigma\) are elementwise power functions, i.e., \(\sigma\left(\cdot\right) = \left(\cdot\right)^p\) for integer values of \(p \geq 2\). Using this assumption, we can apply Theorem K.1 from \citep{kanatsouliscounting} to establish the result. For approximate results, classical smooth nonlinearities such as the hyperbolic tangent, the sigmoid, or the Swish function can be employed and analyzed via their Taylor series expansion around zero:
\begin{equation}
    \sigma(x) = \sum_{p=0}^{{K-1}} \frac{\sigma^{(p)}(0)}{p!}x^p,
\end{equation}
where \(\sigma^{(p)}\) represents the \(p\)-th derivative of \(\sigma(x)\) evaluated at 0. It is straightforward to observe that elementwise power functions appear in this expansion, enabling approximate cycle counting. In any case, the proposed PEs will contain rich information relevant to cycle counts.

{B-\method~achieves the same level of expressivity as R-\method. Specifically, there exists a parametrization $\Phi$, defined by Eq. (\ref{eq:GNNrec02}), such that B-\method~can count the number of 3-, 4-, 5-, 6-, and 7-node cycles in which each node participates for any given graph. This equivalence holds because the input $\bm{x}$ of R-PEARL incorporates higher-order moments, which are directly related to the input $\bm{I}$ of B-\method~as follows: 

$$\mathbb{E}\left[\bm{x}\circ\bm{x}\circ\dots\circ\bm{x}\right]=\bm{I}\circ\bm{I}\circ\dots\circ\bm{I},$$

and the structural expressivity of R-\method~depends on $\mathbb{E}\left[\bm{x}\circ\bm{x}\circ\dots\circ\bm{x}\right]$ according to \citep{kanatsouliscounting}.}

\section{Stability Analysis}\label{app:stability}
Let $\tilde{\mathcal{G}}$ be a perturbed version of graph $\mathcal{G}$, with GSOs $\tilde{\bm S}$ and ${\bm S}$ respectively. We consider two perturbation models, i.e., additive and relative perturbation:
\begin{align}
    &\text{Additive perturbation model:}\quad \tilde{\bm S}={\bm S}+\bm E\\
    &\text{Relative perturbation model:}\quad \tilde{\bm S}={\bm S}\bm E+\bm S\bm E
\end{align}
To measure the distance between $\tilde{\bm S}$ and $\bm S$, as well as the GNN outputs when the input graphs are perturbed versions of each other we define the distance modulo permutation:
\begin{definition}[Linear operator distance modulo permutation] \citep{gama2020stability} \label{def_operator_distance} 
Given linear operators $\bm A$ and $\tilde{\bm A}$ we define the operator distance modulo permutation as
\begin{align} \label{eqn_def_operator_distance}
   \big\| \bm A - \tilde{\bm A} \big\|_{\mathcal{P}} 
      \ = \   \min_{\bm \Pi} \,
                 \max_{ \bm x : \| \bm x \| = 1 } \,
                     \big\|\bm\Pi^{T}(\bm A \bm x)  - \tilde{\bm A} (\bm\Pi^{T} \bm x)\big\|,
\end{align}
where $\bm\Pi$ is a permutation matrix.
\end{definition}

\subsection{Lipschitz and Integral Lipschitz Filters}
Next, we need to define the notion of Lipschitz and Integral Lipschitz filters. First, we note that graph filters are pointwise operators in the graph frequency domain, i.e.,
\begin{equation}
    \bm H\left(\bm S\right)=\sum_{k=0}^{K-1}h_k\bm S^k=\sum_{k=0}^{K-1}h_k \bm V\Lambda^k\bm V^T=\bm V\left(\sum_{k=0}^{K-1}h_k \bm\Lambda^k\right)\bm V^T.
\end{equation}
We can therefore define the graph frequency response of the filter as:
\begin{equation} \label{eqn:frequency_response}
	h(\lambda)  \ = \ \sum_{k=0}^{{K-1}} h_{k} \lambda^{k} .
\end{equation}

To continue our analysis we define the following filter types.
\begin{definition}[Lipschitz Filter] \citep{gama2020stability} \label{def:lipschitzFilters}
Given a filter $\bm h = \{h_{k}\}_{k=0}^{K-1}$ its frequency response $h(\lambda)$ is given by \eqref{eqn:frequency_response}. We say the filter is Lipschitz if there exists a constant $C>0$ such that for all $\lambda_1$ and $\lambda_2$,
\begin{equation}\label{eqn:lipschitzFilters}
   \big| h(\lambda_2) - h(\lambda_1) \big| 
       \leq C \big| \lambda_2 - \lambda_1 \big|.
\end{equation} \end{definition}

\begin{definition}[Integral Lipschitz Filter] \citep{gama2020stability}\label{def:integralLipschitzFilters}
    Given a filter $\bm h = \{h_{k}\}_{k=0}^{K-1}$ its frequency response $h(\lambda)$ is given by \eqref{eqn:frequency_response}. We say the filter is integral Lipschitz if there exists a constant $C>0$ such that for all $\lambda_1$ and $\lambda_2$,
    \begin{equation}\label{eqn:integralLipschitzFilters}
    | h(\lambda_2) - h(\lambda_1) |
    \ \leq\  C\ \frac{| \lambda_2 - \lambda_1 |}{  |\lambda_1 +  \lambda_2| \,/\, 2  }\, .
    \end{equation} \end{definition}

\subsection{Stability bounds for random and basis PEs}
We can use any stability bounds for GNNs. To see that let:
\begin{equation}
    \left\|  \Phi\left( \mathcal{G},\cdot \right)\left[:,f\right]-\Phi\left( \tilde{\mathcal{G}},\cdot \right)\left[:,f\right] \right\|_{\mathcal{P}}\leq \Gamma
\end{equation}
Then
      \begin{align} \label{eq:PEGNN}
 \left\|  \frac{1}{M}\sum_{m=1}^M\Phi\left( \mathcal{G},\cdot \right)\left[:,f\right]-\frac{1}{M}\sum_{m=1}^M\Phi\left( \tilde{\mathcal{G}},\cdot \right)\left[:,f\right] \right\|_{\mathcal{P}} &\leq  \frac{1}{M}\left\|  \sum_{m=1}^M\Phi\left( \mathcal{G},\cdot \right)\left[:,f\right]-\Phi\left( \tilde{\mathcal{G}},\cdot \right)\left[:,f\right] \right\|_{\mathcal{P}}\nonumber\\&\leq  \frac{1}{M}\sum_{m=1}^M\left\|  \Phi\left( \mathcal{G},\cdot \right)\left[:,f\right]-\Phi\left( \tilde{\mathcal{G}},\cdot \right)\left[:,f\right] \right\|_{\mathcal{P}}\nonumber\\&\leq \Gamma
    \end{align}
Using the previous definitions and Eq. (\ref{eq:PEGNN}) we can now use the analysis in \citep{gama2020stability} to establish the stability of the proposed PEs. 

    \begin{proposition}[Stability to additive perturbations] 
    Let $\tilde{\mathcal{G}}$ be a perturbed version of $\mathcal{G}$ such that $\tilde{\bm S}=\bm S+\bm E$ with $\left\| \bm E \right\| \leq \varepsilon$. Let $\Phi$ be an $L-$layer GNN described by Eq. (\ref{eq:GNNrec02}), where each layer consists of Lipschitz filters with constant $C$. Under assumptions \ref{ass:nonlinearity} and \ref{ass:filter} with $C_{\sigma}=1$ and $\beta=1/F$, it holds that:
    % eqn:genericStabilityConstant
    \begin{equation} 
     \left\|  \frac{1}{M}\sum_{m=1}^M\Phi\left( \mathcal{G},\cdot \right)\left[:,f\right]-\frac{1}{M}\sum_{m=1}^M\Phi\left( \tilde{\mathcal{G}},\cdot \right)\left[:,f\right] \right\| _{\mathcal{P}} \leq \left(1+8\sqrt{N}\right) L \varepsilon + \mathcal{O}(\varepsilon^{2})
    \end{equation}
\end{proposition}

    \begin{proposition}[Stability to relative perturbations] 
    Let $\tilde{\mathcal{G}}$ be a perturbed version of $\mathcal{G}$ such that $\tilde{\bm S}=\bm S+\bm S\bm E+\bm E\bm S$ with $\left\| \bm E \right\|_{\mathcal{P}} \leq \varepsilon$. Let $\Phi$ be an $L-$layer GNN described by Eq. (\ref{eq:GNNrec02}), where each layer consists of Integral Lipschitz filters with constant $C$. Under assumptions \ref{ass:nonlinearity} and \ref{ass:filter} with $C_{\sigma}=1$ and $\beta=1/F$, it holds that:
    % eqn:genericStabilityConstant
    \begin{equation} 
     \left\|  \frac{1}{M}\sum_{m=1}^M\Phi\left( \mathcal{G},\cdot \right)\left[:,f\right]-\frac{1}{M}\sum_{m=1}^M\Phi\left( \tilde{\mathcal{G}},\cdot \right)\left[:,f\right] \right\| _{\mathcal{P}} \leq 2\left(1+8\sqrt{N}\right) L \varepsilon + \mathcal{O}(\varepsilon^{2})
    \end{equation}
\end{proposition}

\section{Spectral filters with Graph Filters}\label{app:SPE}
The suggested implementation in \citep{huangstability} is:
\begin{align}
    \text{SPE}\left(\bm V,\bm\Lambda\right) &= \sum_{n=0}^{N-1}\rho\left(\left[\bm V\text{diag}\left(\alpha_1\left(\bm\Lambda\right)\right)\bm V[n]^T,\dots,\bm V\text{diag}\left(\alpha_M\left(\bm\Lambda\right)\right)\bm V[n]^T\right]\right)\\&=
    \sum_{n=0}^{N-1}\rho\left(\left[\bm V\text{diag}\left(\alpha_1\left(\bm\Lambda\right)\right)\bm V^T\bm e_n,\dots,\bm V\text{diag}\left(\alpha_M\left(\bm\Lambda\right)\right)\bm V^T\bm e_n\right]\right),
\end{align}
where $\rho$ represents multiple GIN layers. If we assume that $\alpha_m$ are analytic element wise functions, then we can take the taylor series expansion and represent $\alpha_m$ as a polynomial. Then SPE can be cast as:
\begin{align}
    \text{SPE}\left(\bm V,\bm\Lambda\right) &=\sum_{n=0}^{N-1}\rho\left(\left[\bm V\text{diag}\left(\sum_{k=0}^{K-1} h_k^1\bm\Lambda^k\right)\bm V^T\bm e_n,\dots,\bm V\text{diag}\left(\sum_{k=0}^{K-1} h_k^M\bm\Lambda^k\right)\bm V^T\bm e_n\right]\right)\\&=
    \sum_{n=0}^{N-1}\rho\left(\left[\bm V\sum_{k=0}^{K-1} h_k^1\bm\Lambda^k\bm V^T\bm e_n,\dots,\bm V\sum_{k=0}^{K-1} h_k^M\bm\Lambda^k\bm V^T\bm e_n\right]\right)
\\&=
    \sum_{n=0}^{N-1}\rho\left(\left[\sum_{k=0}^{K-1} h_k^1\bm V\bm\Lambda^k\bm V^T\bm e_n,\dots,\sum_{k=0}^{K-1} h_k^M\bm V\bm\Lambda^k\bm V^T\bm e_n\right]\right)
\\&=
    \sum_{n=0}^{N-1}\rho\left(\left[\sum_{k=0}^{K-1} h_k^1\bm A^k\bm e_n,\dots,\sum_{k=0}^{K-1} h_k^M\bm A^k\bm e_n\right]\right)=
    \sum_{n=0}^{N-1}\rho\left(\sum_{k=0}^{K-1} \bm A^k\bm e_n\bm h_k^T\right)\label{eq:SPE2BPE}
\end{align}
The expression in Eq. (\ref{eq:SPE2BPE}) coincides with the B-\method~architecture, which concludes our proof.

\section{Implementation Details}\label{appendix-implementation}
All results for the SPE, SignNet, and BasisNet models using only 8 eigenvectors were either sourced from their original papers, when available, or obtained by retraining the original models with 8 eigenvectors corresponding to the 8 largest or smallest eigenvalues.
All other baseline results were sourced from their original papers.
For both the R-\method~and B-\method~models, batch normalization is applied within the $\Phi$ layers. Additionally, when $K > 2$, the output of the first $\Phi$ layer is passed through a shallow MLP consisting of 1 or 2 layers before continuing through the remaining layers. 

For the REDDIT datasets, we use R-\method~with 30 samples and $K=2$, omitting the the first layer described in Eq. (\ref{eq:GNNrec02}). Both SignNet and R-\method~use 4 GIN layers with batch normalization to generate the positional encodings, followed by a base model consisting of 6 additional GIN layers. In R-\method, skip connections are applied across those 4 GIN layers, followed by a linear layer at the end. SignNet uses residual connections instead, and also uses MLP encoders for the eigenvectors, as well as a Set Transformer \cite{lee2019settransformerframeworkattentionbased}. For both models, we use a batch size of 70 and 100 on REDDIT-BINARY and REDDIT-MULTI respectively. 

On the ZINC datasets, R-\method, B-\method, and SPE use a batch size of 128. The base model for each is a 4-layer GINE. Similar to the SPE model, we inject the original positional encoding into every layer by passing it through an MLP and adding it to the layer's input. Notably, our model employs 8 GIN layers with 40 hidden units for $\Phi$, whereas SPE uses an 8-layer GIN with 128 hidden units, in addition to 3 MLPs. For R-\method~we use 50-120 samples and $K=12$, while for B-\method~we use $K=4$. 

For the DrugOOD datasets, R-\method, B-\method, SPE, and SignNet all use 4-layer GINE base models. Both R-\method~and B-\method~use a 3-layer GIN for $\Phi$. To process positional encodings, in addition to a 3-layer GIN, SPE uses 16 3-layer MLPs on the Scaffold and Size splits, while SignNet uses a 3-layer MLP across all splits. At each layer of the base model, all models concatenate the original positional encodings with the input features. In the Assay and Size splits we use R-\method~with $K=14$ and 80 samples, while for the Scaffold split, we use $K=16$ and 200 samples. 

{For the RelBench tasks, both R-\method~and B-\method~models use $K=7$. The R-\method model employs a 5-layer GIN with 40 hidden units and 120 samples. The B-\method model uses either a 5-layer or a 7-layer GIN, depending on the task: 5 layers for post-post-related and 7 layers for user-post-comment, both with the same number of hidden units.
For the SignNet models, we use an 8-layer GIN with batch normalization to generate positional encodings. 
The positional encodings from the models are incorporated as additional node features for each node. The original node features are generated by a Tabular ResNet model, which learns representations over the various node features. These combined features are then fed into the base GNN model.
All models follow the same setup as in RelBench for the base model, which employs a 2-layer ID-GNN. Training is conducted with a batch size of 20.}

\begin{table}[t]
\centering
{\caption{{Estimated runtime per epoch in Hours:Minutes for different models on RelBench.}}
\label{tab:relbench_runtime}
\centering
\scalebox{0.8}{\begin{tabular}{l ccccc}
    \toprule
    \textbf{Task} & \textbf{No PE} & \textbf{SignNet-largest} & \textbf{SignNet-smallest} & \textbf{B-\method} & \textbf{R-\method} \\
    \midrule
    \texttt{user-post-comment} & 00:03  & 00:07 & 01:22 & 00:13 & 00:17 \\
    \texttt{post-post-related} & 00:01 & 00:08 & 00:26 & 00:05 & 00:01 \\
    \bottomrule
\end{tabular}}}
\end{table}

Table {\ref{tab:relbench_runtime} presents the runtime of our end-to-end PE models on the RelBench tasks. Notably, our R-\method and B-\method achieve shorter runtimes compared to SignNet-smallest across both tasks.}

{For the Peptides-struct dataset we use an 8-layer GatedGCN base model for B-$\method$, and a 6-layer GINE model for R-$\method$. We use a 9-layer GIN to generate the positional encodings for both models, with $K=4$ for R$-\method$ and $K=1$ for B-$\method$.}

For our experiments and model training pipeline we follow the codebases of \citep{huangstability} and \citep{limsign}, using Python, PyTorch \cite{paszke2019pytorchimperativestylehighperformance}, and the PyTorch Geometric \cite{fey2019fastgraphrepresentationlearning} libraries. Our code can be found here: \url{https://github.com/codelakepapers/RPE-Framework}.

\section{Additional Experiments}
{\subsection{Experiments on Graph Isomorphism}
We conduct experiments on the Circular Skip Link (CSL) dataset \citep{murphy2019relational} which is the golden standard when it comes to benchmarking GNNs for graph isomorphism \citep{dwivedi2023benchmarking}. CSL contains 150 4-regular graphs, where the edges form a cycle and contain skip-links between nodes. Each graph consists of 41 nodes and 164 edges and belongs to one of 10 classes. Message-passing GNNs with WL-related PEs fail to classify these graphs and classification is completely random. This is due to the inability of the WL algorithm to handle regular graphs.

The proposed \method~architectures, however, have no issue in processing regular graphs and achieve $100\%$ classification accuracy. In particular, let $\Phi$ be a two-layer GNN, where each layer is defined by Eq. (\ref{eq:GNNrec02}) with $K=5,~F_0=F_1=1$, and $\sigma(\cdot)= \texttt{ReLU}(\cdot)$. The generated node PE $\bm P$ is processed by a summation graph pooling function to produce a scalar embedding for each graph. Then, both B-\method~and R-\method~can perfectly classify the CSL graphs with $100\%$ classification accuracy, for any randomly generated trainable weights. This means that B-\method~and R-\method~can perfectly classify the CSL graphs without any training. For example let $\Phi$ consist of two identical layers with parameters $(h_0,h_1,h_2,h_3,h_4) = (0,1,-\frac{1}{2},\frac{1}{3},-\frac{1}{4})$. The output $\bm 1^T\bm P$ of B-\method~is presented in Table \ref{table:CSL}. The output remains the same for all graphs within the same class, and differs distinctly for graphs belonging to different classes. Consequently, perfect classification accuracy can be achieved by feeding the B-\method~encoding into a simple linear classifier or even a linear assignment algorithm.}
\begin{table}[t]

{\caption{{B-\method~PE for every class of the CSL graphs. B-\method~can perfectly classify the CSL graphs with $100\%$ classification accuracy}}
\vspace{-0.2cm}
\label{table:CSL}
% \vskip 0.15in
\begin{center}
\begin{small}
\begin{sc}
\scalebox{0.9}{\begin{tabular}{cccccccccc}
\toprule
\multicolumn{10}{c}{Class}\\
0 & 1 & 2 & 3 & 4 & 5 & 6 & 7 & 8 & 9\\
\midrule
\midrule
 0  & 27351.6 & 8800.2& 25779.9& 20458.4&17197.2&15861.3&24055.6&4106.8&17667.0\\
\bottomrule
\end{tabular}}
\end{sc}
\end{small}
\end{center}}
\vskip -0.1in
\end{table}

{\subsection{Experiments on Peptides-struct Dataset}}

\begin{table}[h]
\centering
\caption{Test MAE on the Peptides-struct dataset. B-\method~achieves the second-best performance, behind Graph ViT, within a $500$k parameter budget. Graph ViT utilizes additional random walk information, while B-\method~should learn this information via training.}
\label{tab:peptides_mae}
\scalebox{0.8}{
\begin{tabular}{lcccccccc}
\toprule
 R-\method & B-\method & GPS & SAN+LapPE & SAN+RWSE & GNN-AK$+$ & SUN & Graph ViT \\
\midrule
% \textbf{\#Parameters} & 489k & 491k & - & - & - & - & - \\
 $0.247_{\pm 0.001}$ & $0.248_{\pm 0.001}$ & $0.252_{\pm 0.001}$ & $0.268_{\pm 0.004}$ & $0.255_{\pm 0.001}$& $0.274_{\pm 0.00}$ & $0.250_{\pm 0.001}$ & $0.245_{\pm 0.002}$ \\
\bottomrule
\end{tabular}}
\end{table}
{We conduct experiments on the Peptides-struct dataset from the Long Range Graph Benchmark dataset \cite{NEURIPS2022_8c3c6668}. This dataset comprises over 15,000 graphs containing more than 2 million nodes in total, with each graph ranging from 8 to 444 nodes. It is designed to evaluate a model's ability to capture long-range interactions. The metric used is Mean Absolute Error (MAE), and the task involves regressing on the 3D structure of peptides to predict properties such as their length. We compare the performance of R-\method~and B-\method against state-of-the-art models within $500$k parameter budget: GPS \cite{rampavsek2022recipe}, SAN \cite{kruezer_2021}, SUN \cite{frasca_2022}, GNN-AK \cite{zhao2022from}, and ViT \cite{he2023generalization}. The results can be found in Table \ref{tab:peptides_mae}.}

{We observe that B-\method~achieves the second-best performance, trailing Graph ViT by only $0.002$. However, it is important to note that Graph ViT leverages additional random walk information, whereas B-\method~must learn this information during training. This distinction can be particularly significant, especially given that both models operate within the same constrained 500k parameter budget.}

\section{Ablation Studies}\label{appendix-additional-exps}
{\subsection{Ablation on different backbone models}}

\begin{table}[h]
\caption{Test MAE of B-$\method$ and R-$\method$ with different backbones within a 500k parameter budget.}
\label{tab:backbones}
\centering
\scalebox{0.8}{\begin{tabular}{lccc}
    \toprule
    \textbf{Model} & \textbf{GINE} & \textbf{GatedGCN} & \textbf{PNA} \\
    \midrule
    B-\method & $0.0679 \pm 0.0026$ & $0.0765 \pm 0.0018$ & $0.0740 \pm 0.0010$ \\
    R-\method & $0.0721 \pm 0.0045$ & $0.0810 \pm 0.0039$ & $0.0946\pm 0.0032$ \\
    \bottomrule
\end{tabular}}
\end{table}

We conduct additional base model ablations for our model, using PNA and GatedGCN in addition to GINE on the ZINC dataset \citep{pna-neurips, gatedgcn_iclr}. Our results are shown in Table \ref{tab:backbones}. We observe that \method~ demonstrates strong performance across various backbones. B-\method~ consistently outperforms equivalent SignNet models with the same backbones, while R-\method~ outperforms SignNet with the GatedGCN backbone \citep{limsign}. All models in the table were kept within the 500k parameter budget.

\subsection{Ablation on $K$}

We also report our results on the ZINC dataset \citep{irwin2012zinc} with alternate values of $K$ for R-\method~and B-\method~. In the case of $K=2$ we omit the first layer described in Eq. (\ref{eq:GNNrec02}), using solely an 8-layer GIN for $\phi$. These results are shown in Table\ref{tab:addedrpe1}. We observe that even with $K=2$ our model outperforms SignNet. Furthermore, even with a low $K$ value of 4, B-\method~outperforms SPE.
\begin{table}[t]
\caption{logP Prediction in ZINC with different R-\method~$K$ values}
\label{tab:addedrpe1}
\centering
  \scalebox{0.8}{\begin{tabular}{lcccccc}
    \toprule
    \textbf{PE Method} & \textbf{\#PEs} & $\textbf{K}$ & \textbf{Test MAE} & \textbf{Training MAE} & \textbf{General. Gap} \\  
    SignNet-8S   & 8 & N/A & $0.1034 \pm 0.0056$ & $0.0418 \pm 0.0101$ & $0.0602 \pm 0.0112$ \\
    SignNet   & Full & N/A & $0.0853 \pm 0.0026$ & $0.0349 \pm 0.0078$ & $0.0502 \pm 0.0103$ \\
    BasisNet-8S   & 8 & N/A & $0.1554 \pm 0.0048$ & $0.0513 \pm 0.0053$ & $0.1042 \pm 0.0063$ \\
    BasisNet   & Full & N/A & $0.1555 \pm 0.0124$ & $0.0684 \pm 0.0202$ & $0.0989 \pm 0.0258$ \\
    SPE-8S   & 8 & N/A & $0.0736 \pm 0.0007$ & \textcolor{red}{$0.0324 \pm 0.0058$} & $0.0413 \pm 0.0057$ \\
    SPE   & Full & N/A & $0.0693 \pm 0.0040$ & \textcolor{blue}{$0.0334 \pm 0.0054$} & $0.0359 \pm 0.0087$ \\
    \midrule
    R-\method (ours)& N/A & 1& $0.0699 \pm 0.002$ & $0.0366 \pm 0.006$ & \textcolor{blue}{$0.0333 \pm 0.007$} \\
    R-\method (ours) & N/A & 2 & $0.0831 \pm 0.005$ & $0.0725 \pm 0.0125$ & \textcolor{red}{$0.0106 \pm 0.008$} \\
    R-\method (ours)& N/A & 12 & \textcolor{blue}{$0.0699 \pm 0.002$} & $0.0366 \pm 0.006$ & $0.0333 \pm 0.007$ \\
     B-\method (ours)   & N/A &1& \textcolor{red}{$0.0644 \pm 0.001$} & \textcolor{red}{$0.0290 \pm 0.003$} & \textcolor{red}{$0.0353 \pm 0.002$} \\
    B-\method (ours)   & N/A & 4 & \textcolor{blue}{$0.0680 \pm 0.0023$} & $0.0381 \pm 0.004$ & $0.0299 \pm 0.0033$ \\
    B-\method (ours)   & N/A & 12 & \textcolor{red}{$0.0676 \pm 0.0016$} & $0.0403 \pm 0.0104$ & \textcolor{blue}{$0.0273 \pm 0.0090$} \\
    \bottomrule
  \end{tabular}}
\end{table}

\subsection{Ablation on number of samples $M$}
\begin{figure}[ht]
    \centering
    \begin{subfigure}[b]{0.45\textwidth}
        \centering
        \includegraphics[width=0.8\linewidth]{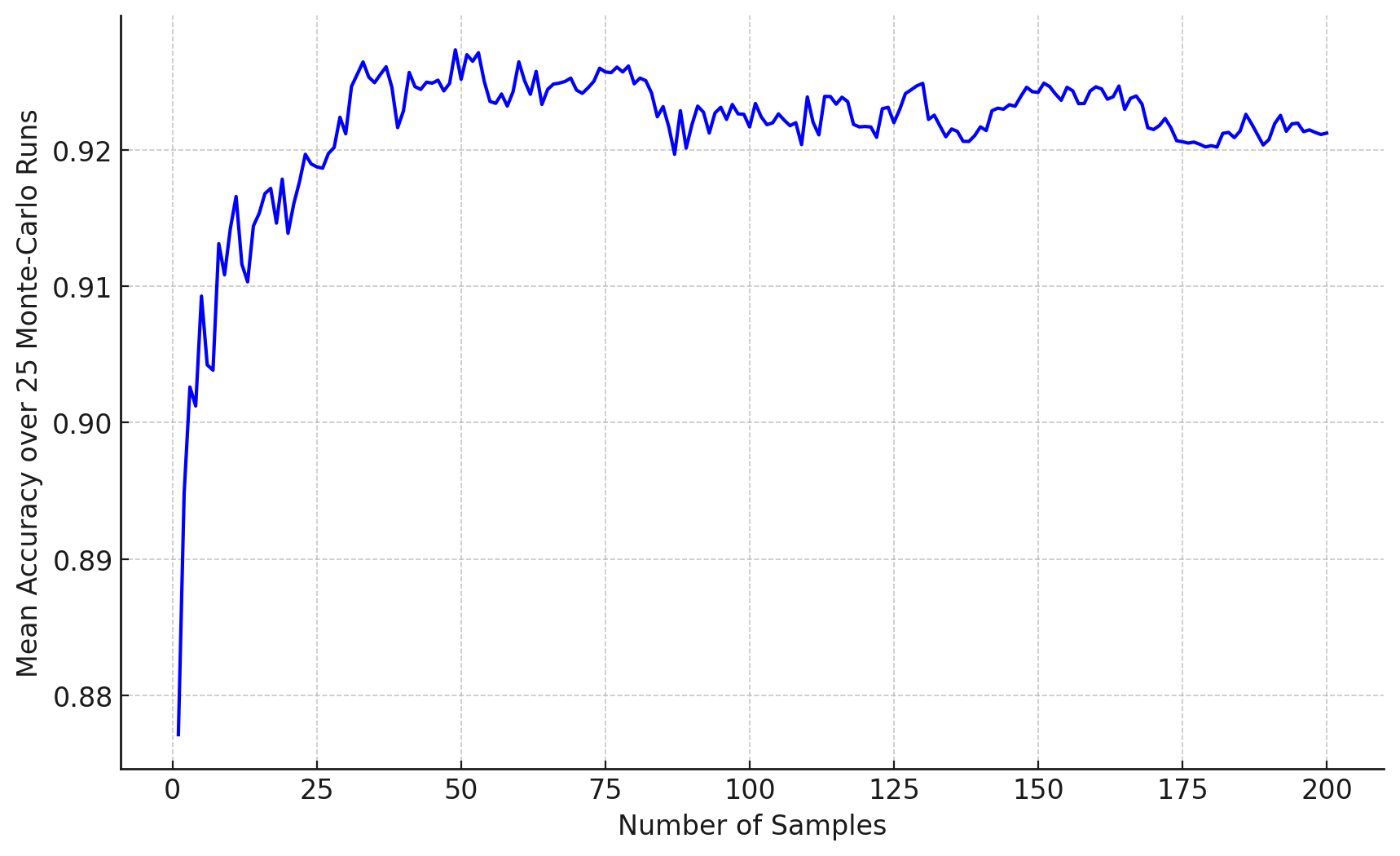}
        \caption{Sample Ablation on REDDIT-BINARY}
        \label{fig:rb-samples}
    \end{subfigure}
    \hfill
    \begin{subfigure}[b]{0.45\textwidth}
        \centering
        \includegraphics[width=0.8\linewidth]{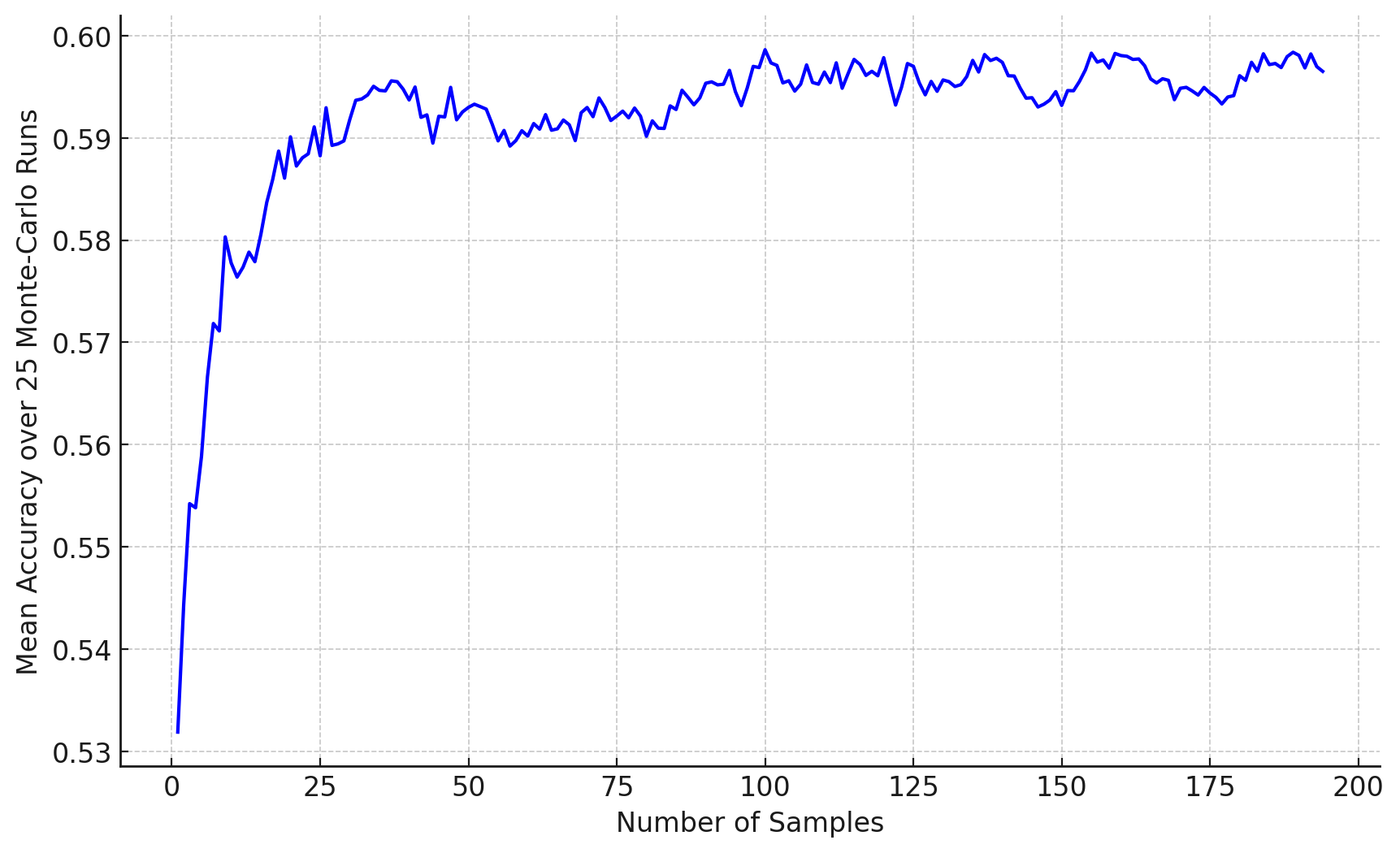}
        \caption{Sample Ablation on REDDIT-MULTI-5K}
        \label{fig:rm-samples}
    \end{subfigure}
    \caption{Ablation studies on the sample size for R-\method; It converges with only a few samples.}
    \label{fig:subgraphs}
\end{figure}

We conduct ablation studies on the number of samples used by R-\method~across the REDDIT datasets to examine the impact of sample size on model performance. These results are illustrated in Fig. \ref{fig:subgraphs}. For each dataset, we evaluate the original model with 200 samples on a single test fold, varying the number of samples from 1 to 200. The test accuracy is then plotted, and Monte Carlo simulation-based smoothing is applied to generate the plots. Notably, we observe that model performance begins to converge with as few as 10 samples—an order of magnitude lower than the graph size.

{\subsection{Ablation on the number of GNN layers in \method}

\begin{table}[ht]
\caption{Ablation on the number of GIN layers in B-\method~for the ZINC Dataset over 4 seeds.}
\label{tab:gin_layers_zinc}
\centering
\scalebox{0.8}{\begin{tabular}{ccccc}
    \toprule
    \textbf{\# GIN Layers} & 3 & 5 & 7 & 9 \\
    \midrule
    \textbf{Test MAE} & $0.0701 \pm 0.005$ & $0.067 \pm 0.003$ & $0.069 \pm 0.001$ & $0.0644 \pm 0.001$ \\
    \bottomrule
\end{tabular}}
\end{table}

The results in Table~\ref{tab:gin_layers_zinc} show that B-$\method$ achieves strong performance on the ZINC dataset even with less layers in the GNN producing the positional encodings. With only 5 layers, B-$\method$ outperforms SPE.
% \clearpage
\subsection{Ablation on different parameter size}
Table \ref{tab:zinc_params} presents our ablation study on the number of parameters for the ZINC dataset. Notably, all models outperform SignNet even within the 500k parameter budget. Furthermore, B-$\method$ achieves superior performance compared to SPE on full eigenvectors, even with fewer parameters.}

\begin{table}[h]
\caption{logP Prediction in ZINC over number of parameters.
}
\label{tab:zinc_params}
\centering
  \scalebox{0.8}{\begin{tabular}{lcccccc}
    \toprule
    \textbf{PE Method} &  \textbf{\#Parameters} & \textbf{Test MAE} & \textbf{Training MAE} & \textbf{General. Gap} \\
    \midrule
    SignNet   & 487k & $0.0853 \pm 0.0026$ & $0.0349 \pm 0.0078$ & $0.0502 \pm 0.0103$ \\
     SPE   & 650k & $0.0693 \pm 0.0040$ & $0.0334 \pm 0.0054$ & $0.0359 \pm 0.0087$ \\
    R-\method & 644k & $0.0699 \pm 0.002$ & $0.0366 \pm 0.006$ & $0.0333 \pm 0.007$ \\
    B-\method & 644k & $0.0644 \pm 0.001$ & $0.0290 \pm 0.003$ & $0.0353 \pm 0.002$ \\
    R-\method & 487k & $0.0721 \pm 0.0045$ & $ 0.0399 \pm 0.001$ & $0.0306 \pm 0.0015$ \\
    B-\method & 487k & $0.0679 \pm 0.0026$ & $0.0336 \pm 0.008$ & $0.0322 \pm 0.004$ \\
    \bottomrule
  \end{tabular}}
\end{table}

\end{document}